\def\paperID{9913}
\def\confName{CVPR}
\def\confYear{2024}

\def\paperTitle{Adapt Before Comparison: A New Perspective on Cross-Domain Few-Shot Segmentation}

\def\authorBlock{
    Jonas Herzog\\
    Zhejiang University \\
    {\tt\small jherzog@zju.edu.cn}
}

\newif\ifreview 
\newif\ifarxiv \newcommand{\arxiv}{\arxivtrue}
\newif\ifcamera 
\newif\ifrebuttal 

\arxiv 

\documentclass[10pt,twocolumn,letterpaper]{article}
\ifreview \usepackage[review]{cvpr} \fi
\ifarxiv \usepackage[pagenumbers]{cvpr} \fi
\ifrebuttal \usepackage[rebuttal]{cvpr} \fi
\ifcamera \usepackage{cvpr} \fi

\usepackage{graphicx}	
\usepackage{algorithm}
\usepackage{algpseudocode}
\usepackage{amsmath}	
\usepackage{amssymb}	
\usepackage{booktabs}
\usepackage{times}
\usepackage{microtype}
\usepackage{epsfig}
\usepackage[table,xcdraw,dvipsnames]{xcolor}
\usepackage{caption}
\usepackage{float}
\usepackage{makecell}
\usepackage{placeins}
\usepackage{color, colortbl}
\usepackage{stfloats}
\usepackage{enumitem}
\usepackage{tabularx}
\usepackage{xstring}
\usepackage{multirow}
\usepackage{xspace}
\usepackage{url}
\usepackage{xcolor}
\usepackage{svg}
\usepackage[hang,flushmargin]{footmisc}

\ifcamera \usepackage[accsupp]{axessibility} \fi

% ...

\ifarxiv  \fi

\newcommand{\R}[1]{{%
    \textbf{%
        \ifstrequal{#1}{1}{\textcolor{red}{R#1}}{%
        \ifstrequal{#1}{2}{\textcolor{blue}{R#1}}{%
        \ifstrequal{#1}{3}{\textcolor{magenta}{R#1}}{%
        \ifstrequal{#1}{4}{\textcolor{teal}{R#1}}{%
                           \textcolor{cyan}{R#1}%
        }}}}%
    }%
}}

\usepackage{xr-hyper}

\makeatletter
\newcommand*{\addFileDependency}[1]{
  \typeout{(#1)}
  \@addtofilelist{#1}
  \IfFileExists{#1}{}{\typeout{No file #1.}}
}

\makeatother

\definecolor{cvprblue}{rgb}{0.21,0.49,0.74}
\usepackage[pagebackref,breaklinks,colorlinks,citecolor=cvprblue]{hyperref}
\usepackage[capitalize]{cleveref}
\crefformat{equation}{Eq. #2#1#3}
\crefname{section}{Sec.}{Secs.}
\crefname{table}{Table}{Tables}
\crefname{figure}{Fig.}{Figs.}

\begin{document}
\title{\paperTitle}
\author{\authorBlock}
\maketitle

\begin{abstract}
Few-shot segmentation performance declines substantially when facing images from a domain different than the training domain, effectively limiting real-world use cases.
To alleviate this, recently cross-domain few-shot segmentation (CD-FSS) has emerged.
Works that address this task mainly attempted to learn segmentation on a source domain in a manner that generalizes across domains.
Surprisingly, we can outperform these approaches while eliminating the training stage and removing their main segmentation network.
We show test-time task-adaption is the key for successful CD-FSS instead.
Task-adaption is achieved by appending small networks to the feature pyramid of a conventionally classification-pretrained backbone.
To avoid overfitting to the few labeled samples in supervised fine-tuning, consistency across augmented views of input images serves as guidance while learning the parameters of the attached layers.
Despite our self-restriction not to use any images other than the few labeled samples at test time, we achieve new state-of-the-art performance in CD-FSS, 
evidencing the need to rethink approaches for the task.
Code is available at \url{https://github.com/Vision-Kek/ABCDFSS}.
\end{abstract}
\section{Introduction}
\label{sec:intro}
With a successful Cross Domain Few Shot Segmentation
(CD-FSS) algorithm, segmentation could be deployed on any task, regardless of the type of objects to segment and its environment.
This paper studies CD-FSS, a task that has emerged recently motivated by the failure of few-shot segmentation (FSS) when test images are fundamentally different from training images.
\begin{figure}[tp]
    \centering
     \includegraphics[width=\linewidth]{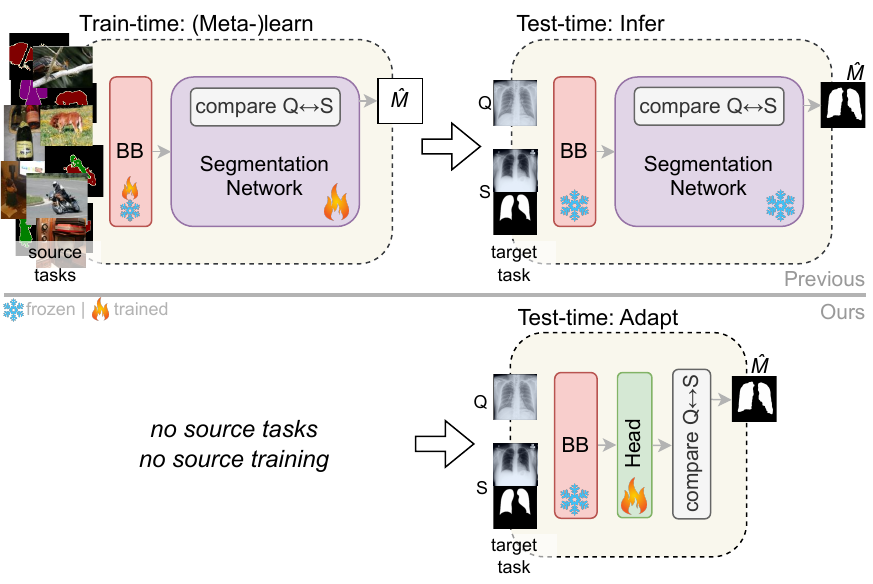}
    \caption{\textit{Top:} Few Shot Segmentation across domains has been addressed by training a deep network on segmentation tasks from a source domain. We demonstrate that its efforts to achieve generalizability during this stage are largely unsuccessful. \textit{Bottom:} In the proposed approach, we entirely forgo such training. Instead, \textit{b}ack\textit{b}one-attached layers (green) adapt features to the target task at test-time.}
    \label{fig:introfig}
\end{figure}

In general, both tasks aim to segment novel classes in a test (query) image based on a few labeled (support) images.
Given this severe knowledge limitation about the novel class, FSS utilizes a base dataset that can provide a larger number of tasks for training.
Since train tasks only provide information about base classes, not the novel classes that will appear at test-time, it is considered crucial that the model can generalize from base to novel classes.

This becomes substantially more challenging when train and test tasks originate from different domains.
Recent approaches for FSS across domains \cite{pat,rtd,dam,hs,repri} focus on this generalization problem and extend FSS with modules designed to enhance knowledge transfer to unseen target domains.
Their learning paradigm and procedure is closely aligned with conventional FSS,
not requiring recent popular large models \cite{segmentanything}. A single source domain such as PASCAL VOC 2012 \cite{pascal} supplies source tasks.
Learning from the source is either conducted by emulating tasks with episodic meta-learning \cite{pat,rtd,dam,hs} or by standard supervised learning \cite{repri}.
A segmentation network is learned on top of a frozen \cite{pat,dam} or trainable \cite{rtd} backbone.
Finally, the model is tested on tasks from the target domain.
The typical architecture and strategy for this is illustrated in Figure \ref{fig:introfig}.
Such approaches rely on similarity-based \textit{comparison} of query and support backbone features in order to locate where the query image matches the support.
Inspired by \cite{dam}, we inspect these similarities.
We find that with a significant domain shift, also lower-level intermediate features are not suitable - the discriminability between semantic classes decreases.
If the representations for a test task are not discriminative, the subsequent segmentation network is predetermined to fail, regardless its generalization ability acquired during train time.
Motivated by this shortcoming, instead of trying to solve the inherently difficult task of learning a generalizable model from a single source domain, we identify \textit{adapting} features to the target task is crucial.

A straightforward solution would be fine-tuning to the test task under utilization of the labeled support set, but it is prone to overfit to the support set \cite{repri,yuhang,huangrestnet}.
\footnote{Such test-time fine-tuning is not to be confused with train-time fine-tuning \cite{svf} and the aforementioned problem to overfit to the base classes \cite{HDM23,hs,dong}, which received the primary attention in FSS research.}
Our solution is a mechanism that relies on embedding consistency within a test task.
Different from all previous work, we do not consider any source tasks.
We demonstrate that adapting ImageNet pretrained backbone features at test-time is sufficient to achieve superior results.
Specifically, we append a small network to each intermediate layer of the backbone.
Both query and support images are augmented to obtain multiple views of them.
Parameters of the attached layers are found as the optimization of a formulation which enforces both class-agnostic embedding consistency and intra-support class consistency across views.
This way we can find features relevant for the current task.
After that, we build query-suppport correlation maps by calculating pixel-to-pixel similarities of the task-adapted features.
The prediction mask is then simply obtained by a parameter-free aggregation of the multi-layer correlation maps.

\begin{itemize}[leftmargin=0.5cm, itemsep=5pt]
    \item Our research reveals that the current approach of learning a downstream FSS network is still inefficient for CD-FSS. We replace it by tiny adaptors that learn at test-time only, proposing Adapt Before Comparison (ABCDFSS).
    \item A novel consistency-based contrastive learning scheme can estimate the parameters of our attached layers without overfitting to the support set. Class discriminability in the query feature space improves significantly. Comparing features from shallow and deep layers separately can then provide domain-shift robust prediction masks.
    \item Our method achieves new state-of-the-art performance on the CD-FSS benchmark and SUIM. Results and experiments highlight the need for our paradigm shift from training a segmentation network to task-adaption.
    \item Our study points out three issues in current CD-FSS work that must be considered in future work: Source domain conceptualization, evaluation metric and benchmark composition.
\end{itemize}
\section{Related Work}
\label{sec:related}
\textbf{Few-Shot Segmentation}
is mainly addressed by comparing the query feature volume with a representation of support foreground class information.
After early approaches with query-support fusion \cite{ca,pfe}, single \cite{dong,sgone} or multiple \cite{pa,pp,asg,pmm} prototypes
became prevalent for the representation of the support class information.
Besides prototype based methods, a more recent branch relies on analyzing pixel-to-pixel correspondences \cite{HDM23,hs,dcama,cwt,pfe}, thus avoiding the loss of spatial structure inherently coming with prototyping.
The base data structure are the dense correspondences between query and support.
Then, either the maximum support correspondence \cite{pfe}, a transformer-style dot product \cite{HDM23,cyctr,dcama,cwt}, or complex learned schemes \cite{hs} are employed to reduce this structure.
While most work resorted to the meta-learning scheme, \cite{cwt} reduced meta-learning to the classifier, \cite{repri} trained the classifier at test-time with no meta-learning and \cite{bam,bam2,HDM23} combined base- and meta-learning branch.
Our test-time learning does not require such strategies.
Self-supervised contrastive learning as in \cite{maxmutual} has been applied for few-shot segmentation in \cite{ganorcon}, its dense variants \cite{densecl,unsuperviseddensecl} have been proposed for large-scale representation learning. 
We use the technique for few-shot task adaption instead.\\
\textbf{Domain Generalization(DG) and Cross-Domain} 
work under domain shifts where no target domain data is accessible, differentiating them from Domain Adaption (DA).
More challenging than DG and DA, in cross-domain few-shot learning (CDFSL) not only the target domain is different from the one seen during training, but also the tasks are novel \cite{cdfs-survey2,cdfs-survey}.
While many previous CDFSL methods \cite{styleadv,dgcdfsl1,dgcdfsl2,rtd,dgcdfsl3} built upon DG techniques to acquire a task-agnostic network in the base step, our paper makes no DG attempts and focuses on adapting to the novel task in the novel domain instead.
Besides fine-tuning \cite{cdfsl20, nsae}, incorporating or attaching small task specific adapters to multiple layers of a deep network has been studied for cross-domain classification \cite{cdfscpr23, litaskadapt} and object detection \cite{acrofod}. Like in our work, these adapters have also been trained from scratch on the target task in \cite{litaskadapt,fsl23}.
Unlike these work for classification, we are interested in dense labels and propose attaching tiny networks that can exploit the dense interaction between support and query.

\textbf{Cross-Domain Few-Shot Segmentation.}\hspace{0.25cm}
A few studies in the FSS literature \cite{fssandbeyond, repri,hs} started evaluating their methods under the small domain shift COCO\cite{coco}\textrightarrow PASCAL\cite{pascal}.
Subsequently, a small number of work focused explicitly on our task, CD-FSS.
RtD \cite{rtd} employs feature enhancement and stores domain-specific style information which is believed to be domain-specific in a memory which is used to generalize during training and guide during testing.
PATNet \cite{pat} prepends a transformation module before HSNet \cite{hs}, leading to more constant prototypes across episodes and domains.
The module is suggested to be fine-tunable on the test task, however, from both design and empirical level the focus is stability at train-time, hence the transformation cannot solve the problem of inadequate features of the target task.
PMNet \cite{dam} proposes a more light-weight architecture based on dense affinity matrices \cite{dcama} between query and support pixels.
One work \cite{yuhang} suggested fine-tuning the backbone on the target task also using the query image, but requires knowledge of target domain unlabeled data. In contrast, we keep the backbone frozen, and assume availability of only one query image as in \cite{rtd,pat,hs,repri}.

Different from all CD-FSS work \cite{rtd,pat,hs,dam,yuhang}, we do not try to learn a domain-generalizing segmentation network. Our method needs no base-, no meta-learning and no source domain data. There are no learnable parameters other than the task-specific weights learned at test-time.
\\

We adopt CD-FSS as the same problem setting as in RtD \cite{rtd} and PATNet \cite{pat}, where access to the target domain is forbidden and classes in the target domain are novel.
\section{Method}
\label{sec:method}

First, following \cite{dcama,hs,pat,pfe}, we use a shared pretrained backbone to extract multi-level features for both support and query images.
Secondly, a small network is appended to each intermediate level of the backbone.
Keeping the backbone frozen, we train this appended network from scratch on the data available at test-time, i.e. the support and query.
Third, given the task adapted features, query pixels that are similar to the support foreground pixels receive a higher foreground score.
A coarse prediction map can be obtained this way for each layer.
Finally, we fuse the layer-wise prediction maps and threshold and optionally refine to obtain the final segmentation mask.

\subsection{Feature Extraction}
\label{subsec:featext}
\newcommand{\fromto}[2]{_{#1}^{#2}}
\newcommand{\feats}[1]{F^{#1}}
\newcommand{\featsL}[1]{\{\feats{#1}_l\}\fromto{l=1}{L}}
\newcommand{\augspecific}[2]{\tilde{{#1}_{#2}}}
\newcommand{\aug}[1]{\augspecific{#1}{a}}
\newcommand{\augfeats}[1]{{\{\feats{\aug{#1}}}\}\fromto{a=1}{aug}}
\newcommand{\smasks}[0]{M^s}

Following \cite{dcama,pat,hs}, query and support images are fed through a pretrained feature extractor to generate multi-layer feature volumes for each.
Due to the structure of the backbone used as feature extractor, the layerwise feature volumes $\feats{q} = \featsL{q}$ and $\feats{s}=\featsL{s}$ have different sized dimensions for different $l$.
Deeper layers, indexed with larger $l$, are smaller in spatial dimensions but larger in the channel dimension.
The support mask is bilinearly downsampled to match the corresponding spatial dimension's size, yielding $\smasks={\{\smasks_l\}}\fromto{l=1}{L}$.

Our method is based on the consistency across views \cite{view} of the same scene.
We augment both query and support geometrically to obtain $AUG$ views of each.
The augmented images are forward passed through the feature extractor in the same way as the original images, resulting in their features ${\{\feats{\aug{q}},\feats{\aug{s}}\}}\fromto{a=1}{AUG}$, where superscripts $\aug{q}$, $\aug{s}$ denote association with the $a$th augmentation of the original image.
After that, we have the augmented features ${\{\feats{\aug{q}},\feats{\aug{s}}\}}\fromto{a=1}{AUG}$ as well as the original $(\feats{q},\feats{s})$.

In our method we want to compare the transformed and original features densely.
Therefore, it is required to maintain the pixel-wise correspondences between original and augmented features.
We restore the correspondences by backprojecting the augmented features with the inverse of the affine that has been applied during augmentation.
For readability it does not receive a new notation.
Only the backprojected augmented features are used in the following.

\begin{figure*}[tp]
    \centering
    \includegraphics[width=\textwidth]{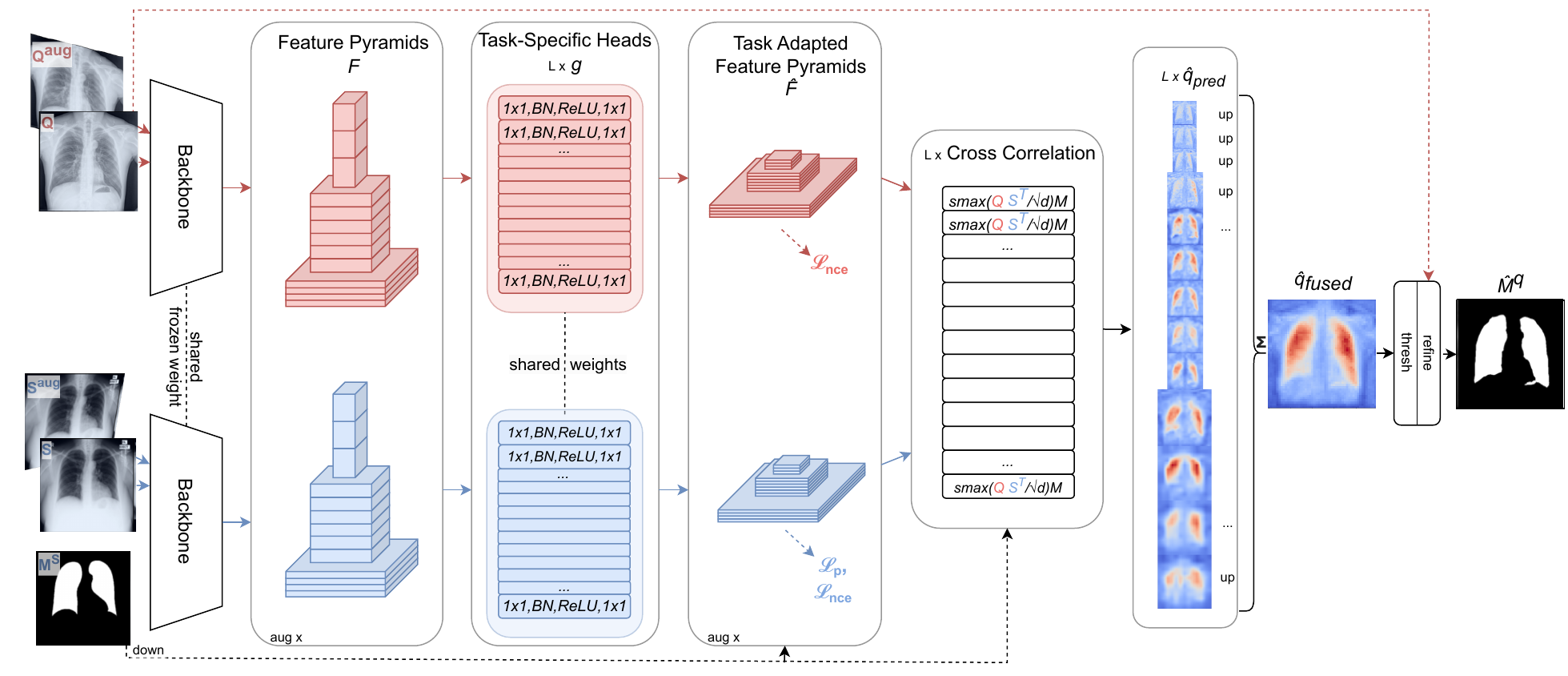}
    \captionsetup{belowskip=0pt, aboveskip=0pt}
    \caption{Overview of proposed method: Query (red) and support (blue) images are augmented to generate views of them. Original image and views are fed separately through a frozen backbone as well as our attached task-specific heads to generate a lower-dimensional feature pyramid. The task-specific networks are trained to maximize intra-level consistency across views. Adapted features are then densely compared in the cross-correlation module. Finally, the level-wise prediction maps are aggregated, thresholded and refined to generate a binary query foreground class prediction.}
    \label{fig:myarch}
\end{figure*}

\subsection{Attached Adapter}
\label{subsec:taskadapt}
Features from the backbone are meaningful in the domain they have been trained on, e.g. ImageNet.
While ImageNet-pretrained weights incorporate a large diversity, for a specific cross-domain few-shot task the embedding space is not optimal.
High intra-class distances and low inter-class distances appear to be prominent issue.

We propose to append a small adapter network to the backbone, specifically one to each of its bottlenecks.
An image can be forward passed through the backbone, yielding $F$, and then through our networks $g=(g_1,g_2,...,g_L$) to obtain task-adapted features $\hat{F}$:
\begin{equation}
\label{eq:taskasdapt}
   \forall l: \hat{F_l} = g_{l}(F_l),
\end{equation}
where $F_l$ represents the intermediate features from the $l-th$ backbone bottleneck.

The small attached networks are trained from scratch on the target query and support set, using self-supervised embedding-alignment and supervised class-alignment.
Training is performed independently for each layer $l$, such that the index $l$ is dropped in the this section readability.
Keep in mind, however, that every term is specific to one network $g_l$.

Reflecting the reduced complexity of the target task compared with ImageNet, the thus generated features are of lower dimensionality, representing distillation of relevant information.

\paragraph{Self-Supervised Embedding Alignment with Dense Contrastive Loss}
We calculate a contrastive loss between features extracted from augmented and non-augmented images, i.e. views.
Dense contrastive learning to match embeddings across views has been proposed for large-scale training \cite{densecl,unsuperviseddensecl}.
Similar to these works, we have a loss term that enforces dot-product similarity of feature volume $\hat{F}$ and its backprojected view $\hat{F}^{aug}$:
\begin{equation}
\label{eq:nce}
    \mathcal{L}_{nce} = \frac{1}{HW}\sum_{i=1}^{HW}{
    -\log\
    \frac{\exp(f_i f_i^{aug}/\tau)}{ \sum_{j=1}^{HW}{\exp(f_i f_j^{aug}/\tau)}}
    },
\end{equation}
where $H,W$ are the spatial dimensions of both $\hat{F}$ and $\hat{F}^{aug}$, from which respective feature vectors $f,f^{aug}$ are extracted using a position index such as $i$ or $j$.
$\tau=0.5$ is a temperature as in \cite{view}.
The enumerator measures the similarity of a positive pair, while the denominator aggregates the similarities of all possible pairs.
Positive pairs are defined as the feature vectors $\in \hat{F}\times \hat{F}^{aug}$ which have the same index.
Complementary, negative pair partners for a vector $f_i\in \hat{F}$ are all vectors $f^{aug}_j\in \hat{F}^{aug},i\neq j$.

For each original image and each of its views, $\mathcal{L}_{nce}$ is calculated independently:
Each of the pairs $(\hat{\feats{q}},\feats{\augspecific{q}{1}}), (\hat{\feats{q}},\feats{\augspecific{q}{2}}),...,(\hat{\feats{q}},\feats{\augspecific{q}{AUG}})$ generates one loss value when plugged into \cref{eq:nce} for ($\hat{F}, \hat{F}^{aug}$).
Equally, multiple $\mathcal{L}_{nce}$ are obtained for the support set.
We average the losses representing embedding discrepancy across views separately for query and support, yielding $L^q_{nce}$ and $L^s_{nce}$.

Complementing the pairwise-correspondence based $L^q_{nce}$ and $L^s_{nce}$, a regularizer is added that acts globally on a feature map.
It ensures consistent statistics and penalizes differences in mean and variance of a feature map $\hat{F}_c$:
\begin{equation}
    \mathcal{L}_{stat} = \frac{1}{C}\sum_{c=1}^{C}{
    \lvert stat(\hat{F}_c) - stat(\hat{F}_c^{aug}) \rvert
    },
\end{equation}
where $C$ is the number of channels and $stat$ yields the statistic of a feature map as a scalar.
This term is calculated for both mean, and variance and then added to the dense contrastive loss, so that we obtain $L^q=\mathcal{L}^q_{nce}+\mathcal{L}^q_{\mu}+\mathcal{L}^q_{var}$ and, by equally but separately calculating $\mathcal{L}_{nce}$ and $\mathcal{L}_{stat}$ for the support features, $L^s$.

\paragraph{Class Alignment with Global Contrastive Loss}
While the previous self-supervised loss does not account for class labels, we introduce the class-aware contrastive prototype loss.
Intuitively, the class prototypes between different views of the same scene should be identical as semantic information is equal.
A contrastive loss motivated by this has been proposed in \cite{rtd}.
In contrast to their usage, our goal is not generalization across domains but generalization \textit{within} the target domain.
Hence, we adopt the term as a support-supervised loss.
The same image pairs and extracted features from \cref{subsec:featext} are used.
The formulation itself is then
\begin{equation}
    \mathcal{L}_{p} = 
    - \log \frac{
    \exp(c(p_{f}, p_{f}^{aug}))
    }{
    \exp(c(p_{f}, p_{f}^{aug}))+\exp(c(p_{f}, p_{b}^{aug}))
    }.
\end{equation}
Foreground prototypes $p_{f}$ and background prototypes $p_{b}$ are obtained by global average pooling \cite{dong} of the respective feature volume 
leveraging the support masks.
Similarities are calculated by cosine similarity  $c(\cdot)$.

Again, we obtain one $\mathcal{L}_p$ for each augmentation, which are subsequently averaged.
For k-shot with $k>1$, prototypes are not calculated $k$ times.
Instead, same-class feature vectors are collectively averaged \cite{dong} when producing class-prototypes.

Since it receives supervision from task-relevant class labels and penalizes low intra-foreground similarity and high inter-class similarity, $\mathcal{L}_p$ should be the main contributor to learn semantically significant features, while the label agnostic self-consistency based $\mathcal{L}^q$ and $\mathcal{L}^s$ constrain the solution space.

For a specific layer $l$ in the pyramid, our attached network $g_l$ is then optimized on the combined loss
\begin{equation}
\mathcal{L} = L^q + L^s + \mathcal{L}_p.
\end{equation}
We observe their contribution to the gradients is balanced and hence not introduce weights.

\subsection{Dense Comparison} 
\label{subsec:comparator}
We calculate the similarity between query and support features to predict the foreground probabilities of query pixels.
With the task-adapted features from \cref{eq:taskasdapt}, measuring similarities between the feature representations $\hat{\feats{q}_l}$ and $\hat{\feats{s}_l}$ is now more semantically meaningful.

We observe dense feature comparison is superior over prototyping for our method and hence adopt the transformer-style query-support-cross-attention weighted mask aggregation from \cite{dcama} to generate a query correlation map $\hat{q}_{pred_l}$ for each layer.
Flattening spatial dimensions in query feature, support features and support masks yields $Q =\hat{\feats{q}_l}, K=\hat{\feats{s}_l}, V=M_l^s$ for
\begin{equation}
\label{eq:damat}
    \hat{q}_{pred_l}=softmax(Q K^T/\sqrt{d})V,
\end{equation}
where $d$ is the size of the channel dimension of $Q$ and $K$, i.e. the dimension over which the dot product is taken.
In \cite{dcama}, positional encoding and a linear projection is used for generating $Q$ and $K^T$ from the feature volumes $F_l^q, F_l^s$ in order to match the transformer architecture.
Because we learned our own head $g_l$ to obtain $\hat{F}_l$ from $F_l$, in \cref{eq:damat} we directly calculate the dot product between adapted query and support feature volumes.
The term does allow $Q$ and $K$ to have different spatial dimensions.
To extend it to k-shot, we follow \cite{dcama} to concatenate support images and masks along the spatial dimension, such that $Q$ is typically of shape $(H\cdot W\times C)$ and $K$ of shape $(H\cdot W \cdot k\times C)$.

\subsection{Segmentation}
Given the layer-wise coarse query prediction maps  $\hat{q}_{pred}$, usually \cite{dcama,hs,dam} a large parametric convolutional downstream segmentation network follows before the final segmentation mask is output.
In our approach, we do not attempt to learn any such.
Instead, we directly fuse the coarse query predictions to obtain a single prediction mask:
\begin{equation}
    \hat{q}_{fused} = \frac{1}{L}\sum_{l=1}^{L}{upsample(\hat{q}_{pred_l})},
\label{eq:fusion}
\end{equation}
where $upsample$ is bilinear interpolation to match the size of the query image.

Because of the softmax from \cref{eq:damat}, resulting maps are in a subrange of $[0,1]$.
The distribution is sample-dependent, however, and cannot be interpreted as probabilities.
Therefore the threshold is chosen such that intra-class variance is minimized, or equivalently, inter-class variance is maximized.
This can be estimated by k-means on the histogram \cite{otsus} of $\hat{q}_{fused}$, such that the binary prediction mask for the query could be obtained as
\begin{equation}
    \hat{M}^q = \hat{q}_{fused} > thresh(\hat{q}_{fused}).
    \label{eq:qmask_nopp}
\end{equation}
Specifically, $thresh$ calculates \cite{otsus}, and if it cannot find a reasonable solution above $mean(\hat{q}_{fused})$, $mean(\hat{q}_{fused})$ is selected as threshold.
See our supplementary for the rationale.

Because of the heavy upsampling ($\times32$ from the highest-level ResNet50 layer), such a prediction mask is only coarse.
A common solution is to skip-connect \cite{dcama,dam} low-level features and then convolve the concatenated features in a decoder to produce a more fine-grained mask.
We observe that using low-level clues in the form of image-appearance and smoothness is sufficient and hence apply \cite{kraehenbuehl} as a non-learnable post-processing to obtain the final prediction mask.

\section{Experiments}
\label{sec:experiments}
\subsection{Experimental Setup}
\begin{figure*}[ht]
  \centering
  \begin{minipage}{.29\linewidth}
    \centering   \captionsetup{belowskip=0pt,aboveskip=4pt,width=.88\linewidth}
    \captionof{table}{FB-IoU is important to report besides mIoU: One could \textit{naive}ly outperform previous SOTA on mIoU by simply assigning foreground to all query pixels (100\%). 1-shot Deepglobe results, true foreground ratio is 43.5\%. \dag: obtained with models trained by ourselves.}
\resizebox{0.9\linewidth}{!}{%
\begin{tabular}[c]{lccc}
\toprule
Method  & mIoU & FB-IoU & \% FG \\
\midrule
Naive & \textbf{43.0} & 21.5 & 100.0 \\
PATNet\dag\cite{pat} & 39.4 & \underline{47.3} & 41.5 \\
Ours & \underline{42.6} & \textbf{47.7} & 48.6 \\
\bottomrule
\end{tabular}
}

    \label{tab:naivedeepglobe}
  \end{minipage}
  \hfill 
  \begin{minipage}{.7\linewidth}
    \centering
    \includegraphics[width=\linewidth]{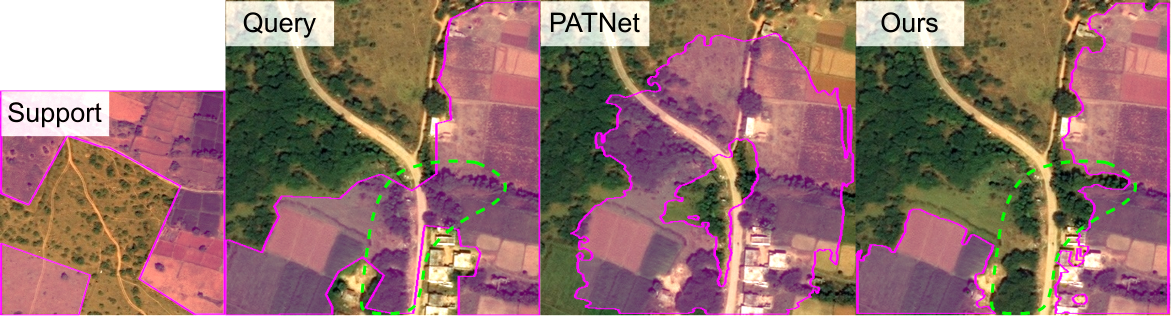}
    \captionsetup{aboveskip=2pt, belowskip=0pt}
    \captionof{figure}{Issue of Deepglobe ground truth annotation. Image row showing an episode featuring the pink overlaid \textit{Agricultural Land} class.
    Green encircled area contains inaccurate inclusion of \textit{Forest} areas in the ground truth (Query) annotation.
    Notably, our model appears to segment agricultural land more precise than the ground truth.}
    \label{fig:deepglobes}
  \end{minipage}
\end{figure*}

\textbf{Metrics.}
Unlike previous CD-FSS, results are reported measuring mIoU \textit{and} FB-IoU.
We argue it is crucial for judging the performance and should always be included.
\cref{tab:naivedeepglobe} shows how previous SOTA could have been outperformed by a naive predictor when only considering mIoU.
The reason is one can get a mIoU boost simply by increasing the predicted foreground ratio.
For definitions, formal derivation and more intuition, please see our Supplementary Material.\\
\textbf{Datasets.}
The primary evaluation datasets are given by the CD-FSS benchmark\cite{pat}.
We align with it and evaluate on Deepglobe\cite{deepglobe}, ISIC\cite{isic}, Chest X-ray (Lung)\cite{chestxray}, FSS-1000\cite{fss1000} in the same way.
Moreover, we compare the results on the underwater dataset SUIM\cite{suim}, following \cite{rtd,dam}.
Unlike these works, there is no source dataset in this work.
As a consequence, PASCAL\cite{pascal} and COCO\cite{coco} have no usage.\\
\textbf{Implementation Details.}
We adopt ResNet50\cite{resnet} with ImageNet\cite{imagenet} pretrained weights as the backbone and follow \cite{hs} to extract features at the end of each bottleneck before ReLU, resulting in 13 layers.
Two augmentations per image are generated, using random shearing of maximum absolute 20 degrees.
Similar to \cite{repri}, our layers are trained with SGD for 25 epochs with learning rate $0.01$.
Each task adaptor is equally defined as $g_l=Conv_{1\times1}(ReLU(BN(Conv_{1\times1}(F)))$ where the $1\times1$ convolutions have 64 output channels.
For postprocessing, we set standard deviations to 1 for spatial Gaussian, 35 for spatial bilateral, 13 for color as well as compatibilities of 2 and 1 for Gaussian and bilateral, respectively and apply it only if it can increase the intersection over union of a pseudoepisode, where the support image functions as pseudoquery and its augments as the pseudosupport.
\subsection{Comparison with State-of-the-art}
\begin{table*}
\captionsetup{aboveskip=4pt, belowskip=0pt}
\caption{Results and comparison on the CDFSS benchmark\cite{pat} on mIoU.
PMNet\cite{dam} has not reported class-wise ISIC results.}
\label{tab:benchmark}
\resizebox{\textwidth}{!}{
\begin{tabular}{lccccccccccc}
\toprule
& \multicolumn{5}{c}{1-shot} & \multicolumn{5}{c}{5-shot}\\
\cmidrule(r){2-6} \cmidrule(r){7-11}

Method  & Deepglobe & ISIC &  X-ray & FSS-1000 & \textbf{Avg.} & Deepglobe & ISIC & X-ray & FSS-1000 & \textbf{Avg.} \\
\midrule
$Linear_{ResNet}$ & 34.1 & 20.8 & 59.1 & 41.0 & 38.8 & 46.5 & 34.8 & 64.6 & 58.7 & 51.1\\
$Linear_{Deeplab}$ \cite{pat} & 33.0 & 19.4 & 43.5 & 40.5 & 34.1 & 39.7 & 30.0 & 60.3 & 58.4 & 47.1\\
PANet\cite{pa}\small\textcolor{gray}{(ECCV20)} & 36.6 & 25.3 & 57.8 & 69.2 & 47.2 & \underline{45.4} & 34.0 & 69.3 & 71.7 & 55.1 \\
RePRI\cite{repri}\small\textcolor{gray}{(CVPR21)} & 25.0 & 23.3 & 65.1 & 71.0 & 46.1 & 27.4 & 26.2 & 65.5 & 74.2 & 48.3 \\
HSNet\cite{hs} {\small\textcolor{gray}{(ICCV21)}} & 29.7 & 31.2 & 51.9 & 77.5 & 47.6 & 35.1 & 35.1 & 54.4 & 81.0 & 51.4 \\
PATNet\cite{pat}\small\textcolor{gray}{(ECCV22)} & \underline{37.9} & 41.2 & 66.6 & 78.6 & \underline{56.1} & 43.0 & \textbf{53.6} & 70.2 & 81.2 & \underline{62.0} \\
HDMNet\cite{HDM23}\small\textcolor{gray}{(CVPR23)} & 25.4 & 33.0 & 30.6 & 75.1 & 41.0 & 39.1 & 35.0 & 31.3 & 78.6 & 46.0 \\
RestNet\cite{huangrestnet}\small\textcolor{gray}{(BMVC23)} & 22.7 & \underline{42.3} & 70.4 & \underline{81.5} & 54.2 & 29.9 & 51.1 & 73.7 & \underline{84.9} & 59.9 \\
PMNet\cite{dam}\small\textcolor{gray}{(WACV24)} & 37.1 & - & \underline{70.4} & \textbf{84.6} & - & 41.6 & - & \underline{74.0} & \textbf{86.3} & - \\

ABCDFSS (Ours) & \textbf{42.6} & \textbf{45.7} & \textbf{79.8} & 74.6 & \textbf{60.7} & \textbf{49.0} & \underline{53.3} & \textbf{81.4} & 76.2 & \textbf{65.0} \\
\bottomrule
\end{tabular}
}
\end{table*}

\begin{table}
\captionsetup{aboveskip=0pt, belowskip=0pt}
\caption{Results and comparison on SUIM, 1-shot, mIoU.
}
\resizebox{\linewidth}{!}{%
\begin{tabular}{lcccccc}
\toprule
HS\cite{hs} & SCL\cite{scl} & RtD\cite{rtd} & Rest\cite{huangrestnet} & PM\cite{dam} & Ours\\
\midrule
28.8 & 31.8 & 34.7 & 25.2 & 34.8 & \textbf{35.1}\\
\bottomrule
\end{tabular}
}
\label{tab:suim}
\end{table}

All previous work in our comparisons is trained on PASCAL VOC 2012\cite{pascal} or, for \cite{HDM23}, on the even richer COCO\cite{coco}.
Our method has seen no dataset.
\vspace{0.05cm}\\
Table \ref{tab:benchmark} compares our work on the CD-FSS benchmark \cite{pat}.
In both 1-shot and 5-shot, we surpass PMNet on mIoU by significant margins of 5.5, 7.4 on Deepglobe and by 8.6,7.4 on Chest X-ray, while underperforming on FSS also by a large 10.0 and 10.1.
FSS underperformance is due to the character of our approach which does not attempt to learn a segmentation network.
As a consequence we can find local semantic similarity well, but a) not learn global semantic clues as a segmentation network's encoder would, and b) not learn spatial accuracy as a segmentation network's decoder would.
This impacts our performance on FSS, where finding the object is generally easy, and performance is gained by spatial accuracy.
For ISIC, PMNet\cite{dam} treats all images as if they belonged to the same semantic class.
Hence, they report mIoU by only ``averaging'' the IoU of one class, which forbids comparison with the CD-FSS benchmark.
Nevertheless, we also compare with their ISIC setting, where our work performs better with 51.3(+0.2) on 1-shot and 59.2(+4.7) on 5-shot.
Importantly, the CD-FSS benchmark average from previous SOTA\cite{pat} is surpassed by our method on average by 4.6 on 1-shot and 3.0 for 5-shot.
\cref{tab:suim} shows our method can also surpass PMNet\cite{dam} on SUIM by 0.3 against a 0.1 of PM over second-best RtD\cite{rtd}.\\
In accordance with our findings that FB-IoU needs to be considered as well, \cref{tab:fullours} reports our full results compared with PATNet\cite{pat} trained by ourselves and the recent FSS-method HDMNet\cite{HDM23}.
Instable validation curves during training \cite{pat} are observed, causing variations from their reported results, but the overall trend remains stable.
HDMNet results are obtained from the meta-branch mask as it was more accurate than the fused mask.
Table \ref{tab:fullours} further documents that even without refinement our results can surpass previous work.
\begin{table*}[ht]
\centering
\captionsetup{aboveskip=4pt,belowskip=0pt}
\caption{Our complete results. \textit{m}IoU and \textit{FB}-IoU.
\textit{No-pp} reports the performance of the unrefined prediction $\hat{M}^q$ from \cref{eq:qmask_nopp}. \dag: Results obtained with models trained by ourselves.}
\resizebox{\textwidth}{!}{%
\begin{tabular}{lcccccccccccccccccccccccc}
\toprule
\multirow{3}{*}{Method} & \multicolumn{4}{c}{Deepglobe} & \multicolumn{4}{c}{ISIC} & \multicolumn{4}{c}{X-ray} & \multicolumn{4}{c}{FSS-1000} & \multicolumn{4}{c}{\textbf{CD-FSS Avg.}} & \multicolumn{4}{c}{SUIM} \\
& \multicolumn{2}{c}{1-shot} & \multicolumn{2}{c}{5-shot} & \multicolumn{2}{c}{1-shot} & \multicolumn{2}{c}{5-shot} & \multicolumn{2}{c}{1-shot} & \multicolumn{2}{c}{5-shot} & \multicolumn{2}{c}{1-shot} & \multicolumn{2}{c}{5-shot} & \multicolumn{2}{c}{1-shot} & \multicolumn{2}{c}{5-shot} & \multicolumn{2}{c}{1-shot} & \multicolumn{2}{c}{5-shot} \\
\cmidrule(lr){2-3} \cmidrule(lr){4-5} \cmidrule(lr){6-7} \cmidrule(lr){8-9} \cmidrule(lr){10-11} \cmidrule(lr){12-13} \cmidrule(lr){14-15} \cmidrule(lr){16-17} \cmidrule(lr){18-19} \cmidrule(lr){20-21} \cmidrule(lr){22-23} \cmidrule(lr){24-25}
& m & FB & m & FB & m & FB & m & FB & m & FB & m & FB & m & FB & m & FB & m & FB & m & FB & m & FB & m & FB \\
\midrule
PATNet\dag\cite{pat} & 35.4 & 45.7 & 41.6 & 50.2 & 43.4 & 62.1 & 51.8 & 68.8 & 63.0 & 72.6 & 63.9 & 73.3 & 77.7 & 85.5 & 79.8 & 87.2 & 54.9 & 66.5 & 59.3 & 69.9 & 32.1 & \textbf{54.2} & 40.2 & 57.8 \\
HDMNet\cite{HDM23} & 25.4 & 38.4 & 39.1 & 46.4 & 33.0 & 49.7 & 35.0 & 50.4 & 30.6 & 25.8 & 31.3 & 28.8 & 75.1 & 84.1 & 78.7 & 86.5 & 41.0 & 49.5 & 46.0 & 53.0 & 23.4 & 49.5 & 30.9 & 51.5 \\
\textbf{Ours (no-pp)} & 42.3 & 47.1 & 48.2 & 53.4 & 41.8 & 57.2 & 50.8 & 63.9 & 80.0 & 86.2 & 81.6 & 87.4 & 69.3 & 79.3 & 73.1 & 82.3 & \underline{58.3} & \underline{67.5} & \underline{63.4} & \underline{71.8} & \underline{35.0} & \textbf{54.2} & \underline{41.1} & \textbf{58.3} \\
\textbf{Ours} & 42.6 & 47.7 & 49.0 & 54.6 & 45.7 & 60.3 & 53.3 & 66.1 & 79.8 & 86.1 & 81.4 & 87.3 & 74.6 & 82.7 & 76.2 & 84.2 & \textbf{60.7} & \textbf{69.2} & \textbf{65.0} & \textbf{73.1} & \textbf{35.1} & 53.5 & \textbf{41.3} & \underline{58.2} \\

\bottomrule
\end{tabular}
}
\label{tab:fullours}
\end{table*}

\subsection{Computational Efficiency}
\begin{table}
  \centering
    \captionsetup{belowskip=0pt,aboveskip=4pt}
  \caption{Task-adapting to a single 1-shot episode and subsequently forward passing other queries. Performance gap to fitting for every episode is reported.}
  \label{tab:yoto}
  \resizebox{\linewidth}{!}{%
  \begin{tabular}{clcccccc} 
  \toprule
    && Deepgl. & ISIC & X-ray & FSS & SUIM & Avg. \\
    \midrule
    \multirow{2}{*}{quick-infer} & \multicolumn{1}{l}{mIoU} & -0.01 & 0.01 & -0.03 & -0.71 & -0.01 & -0.15 \\
    & \multicolumn{1}{l}{FB-IoU} & -1.08 & -0.79 & -0.29 & 0.01 & -2.87 & -1.00 \\ 
  \bottomrule
  \end{tabular}
  }
\end{table}

For a comparison with previous work under equal conditions, the method has been evaluated on episodes consisting of only one query image.
In reality, multiple queries might want to be processed subsequently.
Since we do test-time adaption utilizing the query, rerunning the task-adaption for every query can be too slow for some application cases.
We therefore evaluate the scenario where task-adaption is only done once, and the thus learned parameters are reutilized for every subsequent query.
Under this setting the performance of our method remains stable, with a maximum mIoU drop of 0.7 on FSS-1000 and maximum FB-IoU drop of 2.9 on SUIM, while cutting computational cost to $\sim1/50$.
This highlights that the task-adaption on one query-support pair is able to generalize to other queries.
For further speedup, unrefined results are compared.
Table \ref{tab:yoto} reports the results averaged over 200 runs, where a run samples first an episode for training and then infers on further 200 queries.

\subsection{Architectural Validation and Ablation}
\label{subsec:ablation}
\textbf{Loss Terms.}
\begin{table}
  \centering
  \captionsetup{aboveskip=4pt,belowskip=0pt}
  \caption{1-shot performance drop when leaving one loss term out or when using ResNet layers directly as an input for the dense comparison without task-adaption (TA).
  }
  \label{tab:leavelossout}
  \resizebox{\linewidth}{!}{%
  \begin{tabular}{clcccccc}
  \toprule
    && Deepgl. & ISIC & X-ray & FSS & SUIM & Avg. \\
    \midrule
    \multirow{2}{*}{w/o $L^q,L^s$} & \multicolumn{1}{l}{mIoU} & -0.06 & -8.91 & -1.86 & -0.18 & 0.69 & -2.06 \\
     & \multicolumn{1}{l}{FB-IoU} & 0.31 & -6.75 & -1.02 & 0.13 & 1.27 & -1.21 \\
    \midrule
    \multirow{2}{*}{w/o $L_{p}$} & \multicolumn{1}{l}{mIoU} & -1.54 & -6.52 & -4.55 & -1.44 & -0.95 & -3.00 \\
     & \multicolumn{1}{l}{FB-IoU} & -1.52 & -6.81 & -3.45 & -0.66 & -1.48 & -2.78 \\
    \midrule
    \multirow{2}{*}{w/o TA} & \multicolumn{1}{l}{mIoU} & -3.33 & -7.12 & -0.03 & -15.72 & -3.09 & -5.86 \\
     & \multicolumn{1}{l}{FB-IoU} & -0.87& -5.68 & -0.06 & -10.31 & 0.92 & -3.19 \\
     \bottomrule
  \end{tabular}
  }
\end{table}

\begin{figure}
    \centering
    \captionsetup{belowskip=0pt, aboveskip=4pt}
     \includegraphics[width=\linewidth]{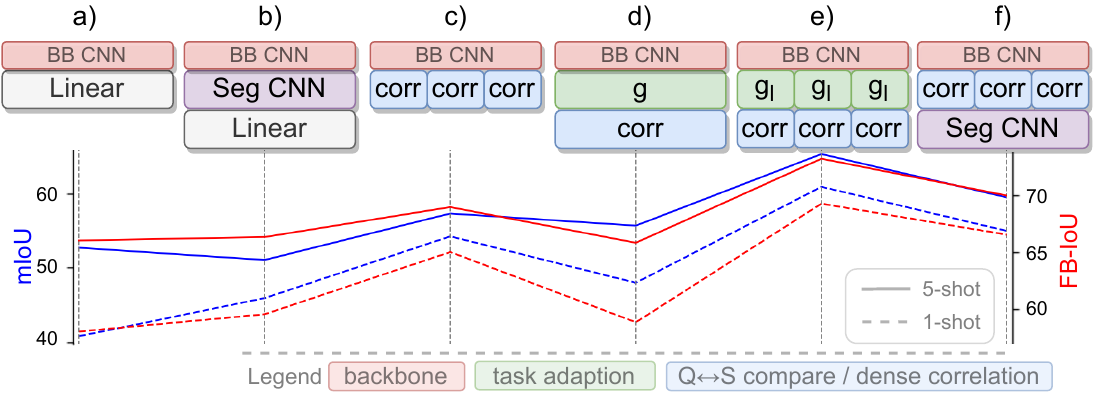}
     \caption{Adapt before comparison e) is the superior approach for CD-FSS. 
     Architectural schemes and their CD-FSS Avg. performance: a) $Linear_{ResNet}$, b) transductive FT \cite{repri}, c) w/o TA, d) hypercolumn TA, e) proposed f) prev. SOTA \cite{pat}, also \cite{dam,HDM23,hs,dcama}.
     }
     \label{fig:archschemes}
\end{figure}

As shown in \cref{tab:leavelossout}, both the unsupervised $L^{q},L^{s}$ and support-supervised $L_p$ contribute to performance enhancement.
Interestingly, for FSS-1000, either of them would be sufficient, while for others such as ISIC, only using one term would be harmful.
While task adaption is beneficial in all scenarios (compare \cref{fig:archschemes} c) and \cref{tab:leavelossout}), the gap for X-ray is surprisingly small, given that previous work \cite{litaskadapt,dam,pat} all considered it to have a large domain shift.
It implies that intermediate features from the backbone would here be already sufficient.
In contrast to the previous segmentation networks which are more harmful (at least -7.4 mIoU) than useful, our multi-layer similarity score aggregation from \cref{eq:fusion} proves here to preserve discriminability: Maps with higher confidences receive implicitly higher scores and thus higher weights in the subsequent summation.

\begin{figure}[ht]
  \centering
  \begin{minipage}{\linewidth}
    \centering
    \captionsetup{aboveskip=4pt, belowskip=0pt}
    \captionof{table}{Intra- and inter-class similarities in the embedding space of (L)ow, (M)iddle and (H)igh-level feature maps before and after \textit{T}ask \textit{A}daption. Measure represents averaged cosine similarities of pixel pairs from same and opposite classes, respectively. \textit{Across} SUIM dataset images. A higher delta represents higher discriminability. Full table on all datasets with more extensive measures can be found in supplementary.}
    
\resizebox{0.75\linewidth}{!}{%
\begin{tabular}{l|ccc|ccc} 
\midrule
Similarities across & \multicolumn{3}{c|}{w/o TA} & \multicolumn{3}{c}{with TA} \\ 
image pairs $\cdot100$ & L & M & H & L & M & H \\ 
 
\midrule
\texttt{FG$\leftrightarrow$FG (INTRA)} & 51 & 42 & 58 & 14 & 20 & 16 \\
\texttt{FG$\leftrightarrow$BG (INTER)} & 50 & 41 & 57 & -4 & -4 & -3 \\
\textbf{\textit{delta}} & \textbf{1} & \textbf{2} & \textbf{1} & \textbf{18} & \textbf{24} & \textbf{18} \\
\end{tabular}
}
    \label{tab:embeddings}
  \end{minipage}
  \begin{minipage}{\linewidth}
    \centering
    \includegraphics[width=\linewidth]{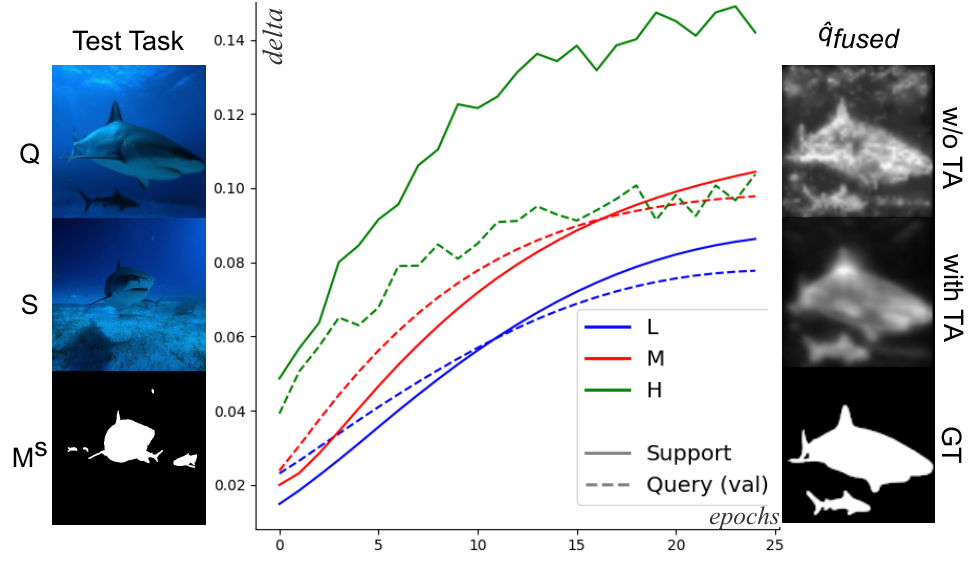}
    \captionsetup{aboveskip=0pt, belowskip=0pt}
    \captionof{figure}{Against common belief, fine-tuning does not lead to overfitting to the support set with our approach. Through learning of consistent embedding spaces, we enhance class discriminability not only for the support (solid lines), but also for the test query (dashed). As a result, irrelevant regions are no longer activated in the coarse query prediction \textit{with TA}.}
    \label{fig:curves}
  \end{minipage}
  \label{fig:taskadaptminipage}
\end{figure}
\textbf{Discriminabilty in Embedding Space.}
Lei \etal \cite{pat} also identify the class distinction as a primary issue for CD-FSS, but measure it using final features of Inception \cite{inception} network.
However, it is multi-layer ResNet features that are relevant for both their and our network.
To obtain a more relevant metric, we reconstruct the dense affinity\cite{dam} matrix which is the dot product part of \cref{eq:damat}.
We measure intra-support through constructing $S S^T$ and across-image through $Q S^T$, where $Q$ and $S$ represent features from a sampled query and support image.
Intra-support metrics show how well the model is fit to the relevant class given a single image.
Across-image metrics are crucial for generalization from support to query.
\cref{fig:curves} and \cref{tab:embeddings} demonstrate the underlying issue of near-zero across-image discriminablilty found initially and how it improves substantially during the learning of our attached layers.
This way, the core precondition for the subsequent comparison is restored.\\
\textbf{Against other fine-tuning or transfer learning.}
We validate our proposed architecture by comparing against alternative approaches as shown in \cref{fig:archschemes}.
In \cite{ganorcon}, few-shot segmentation is addressed by appending a small projector network to hypercolumns from a contrastively pretrained encoder.
A hypercolumn is the concatenation of features from $L$ layers to shape \begin{small}$H\times W\times(C_0+C_1+...+C_{L-1})$\end{small}, with upsampling of higher-level features before concatenation, if necessary.
This is a viable alternative idea for our level-wise approach.
In the \textit{first setting} $Linear_{ResNet}$, \cref{fig:archschemes} a), we compare with training a linear classifier on the support set, thus mapping backbone hypercolumn directly to foreground probability.
Average benchmark performance decreases by mIoU\textbar FBIoU ($-19.6|-11.1$) for 1-shot and ($-12.3|-7.1$) for 5-shot, proving that simple fine-tuning or transfer learning cannot compete with our method. Compare also $Linear_{Deeplab}$ in \cref{tab:benchmark} as well as transductive fine-tuning \cref{fig:archschemes} b).
In the \textit{second setting}, \cref{fig:archschemes} d), we replace our $L$ task adapters with a \textit{single task-adapter network} which takes the ResNet-extracted hypercolumn as input and produces a single feature representation per image.
Given query and support representations, the dense comparison module from \cref{subsec:comparator} generates the similarity prediction.
No subsequent fusion is required since there is only one prediction map.
Average benchmark performance loss is m\textbar FB $(-12.6|-10.3), (-9.4|-7.3)$ for 1- and 5-shot respectively,
which suggests that mixing the information from different levels is not as generalizable as comparing them individually.
Instead, the simple averaging for fusion proves to be effective for self-regularization and noise suppression.

\section{Discussion}
\label{sec:discussion}
\textbf{Benchmark and Datasets}\hspace{0.25cm}
In addition to the demonstrated need to complement mIoU with FB-IoU, there are a few more points to consider.
First, we agree with the suggestion \cite{dam} to differentiate between \textit{cross-dataset} and \textit{cross-domain} few-shot segmentation.
FSS-1000\cite{fss1000} from the CD-FSS benchmark is classified as \textit{cross-dataset} and is hence useful to understand performance in domains where conventional FSS methods also perform well, but the underwater-dataset SUIM\cite{suim} used in \cite{rtd} is more appropriate to consider for pure CD-FSS.
Second, the benchmark includes Deepglobe, specifically \cite{pat} the Land Cover Classification Dataset.
\cref{fig:deepglobes} illustrates that this dataset is improperly annotated, limiting the expressiveness of performance measure.
Even though we outperform previous work on Deepglobe, we suggest to find a properly annotated sattelite image segmentation alternative for future work.

\textbf{Source Domain}\hspace{0.25cm}
Our method does not use any dataset for training, yet we refrain from calling it zero-source or source-free to avoid confusion. ImageNet is the source domain of the pretrained backbone. While previous CD-FSS research also uses ImageNet weights, they declare only \textit{their} training domain, i.e. PASCAL or COCO, as the source domain.
It might be technically valid since the segmentation network does not see images from ImageNet. However, they suggest misleadingly that the primary challenge lies in bridging the domain gap between e.g. PASCAL and SUIM, neglecting the pivotal role of ImageNet-based features in semantic information transfer. Our results, powered by the backbone features alone, underscore the necessity to acknowledge this across related works.

\textbf{Limitations and Future Work}\hspace{0.25cm}
By removing the learning of a segmentation network, we intend to expose the fundamental problem in cross-domain few-shot segmentation and show that it should be addressed by test-time task adaption. However, we do not see our method as a final solution for CD-FSS.
As it adapts features vector-wise, it does not learn the scene-level semantic context which is commonly seen as a key for semantic segmentation. While it is out of scope of this work, we believe that if our findings are addressed appropriately, replacing the heuristic fusion and refinement through reintroducing training on source segmentation tasks could improve performance in future work.

\section{Conclusion}
\label{sec:conclusion}
Previous sophisticated similarity fusion models are not yet effective for CD-FSS.
We presented a cross-domain few-shot segmentation method that outperforms previous approaches with no segmentation network.
Averaging similarities across a feature pyramid is simpler and more effective, provided that the features are task-adapted before calculating their similarities.
Enforcing contrastive consistency proved to be a strategy that could avoid overfitting to the support set while fine-tuning.
Results and experiments suggest task-adaption before comparison is the superior approach for CD-FSS.
{\small
\bibliographystyle{ieeenat_fullname}
\bibliography{11_references, refs/seg, refs/fsl, refs/fss, refs/cd, refs/cdfss, refs/datasets,
refs/lossterms}
}

\ifarxiv \clearpage \appendix \section{Metrics}
We claimed reporting mIoU only does not reflect performance appropriately.
The relationship between mIoU, FB-IoU and foreground ratio is derived in the following.

\paragraph{mIoU}
Mean intersection over union (mIoU) is calculated by
\begin{enumerate}
    \item Accumulation of intersection areas, union areas over query predictions $q$  \cref{eq:ic_uc}
    \item Calculating IoU for each semantic class,\cref{eq:iou_c}
    \item Averaging the class-wise IoUs, \cref{eq:miou}
\end{enumerate}

\begin{equation}
\begin{gathered}  
I_c=\sum_q{TP_{q,c}}\\
U_c=\sum_q{TP_{q,c}+FP_{q,c}+FN_{q,c}},
\end{gathered}
\label{eq:ic_uc}
\end{equation}

\begin{equation}
IoU_c = \frac{I_c}{U_c}
\label{eq:iou_c}
\end{equation}
\begin{equation}
mIoU = \frac{1}{C} \sum_{c=1}^{C}IoU_c
\label{eq:miou}
\end{equation}

with true positives $TP$ counting pixels where both prediction and ground truth label equal $c$, $FP$ being the number of pixels where $c$ was falsely predicted and $FN$ the amount of ground truth $c$-labels which were predicted as another class.

Note that the number $C$ of categories in the dataset does not include the background class.


\paragraph{FB-IoU}
In 1-way segmentation, which we and all previous CD-FSS focus on, the task is binary segmentation: For each episode, a class $c$ is selected, query and support containing $c$ are sampled, $c$ is treated as \textit{f}oreground and everything $\neq c$ is treated as complementary \textit{b}ackground.
Foreground background intersection over union is calculated through
\begin{enumerate}
    \item Accumulating the areas of intersection and union with respect to both $c$ and $\neq c$ - \cref{eq:ic_uc,eq:ineqc_uneqc}
    \item Treating all classes equal by aggregating their metrics to fore-and background - \cref{eq:if_bf}
    \item Averaging IoU for foreground and background - \cref{eq:iou_fb,eq:fbiou}
\end{enumerate}

We can obtain the background class metrics in the style of \cref{eq:ic_uc} through
\begin{equation}
\begin{gathered}
    I_{\neq c} = \sum_q{TN_{q,c}}\\
    U_{\neq c} = \sum_q{TN_{q,c}+FN_{q,c}+FP_{q,c}},
\end{gathered}
\label{eq:ineqc_uneqc}
\end{equation}
where $TN_{q,c}$ indicates that both prediction and ground truth did not predict $c$.
The foreground and background intersections and unions are then obtained through
\begin{equation}
\begin{gathered}
    I_f = \sum_c I_c,\hspace{0.5cm}
    I_b = \sum_c I_{\neq c} \\
    U_f = \sum_c U_c,\hspace{0.5cm}
    U_b = \sum_c U_{\neq c}
\end{gathered}
\label{eq:if_bf}
\end{equation}
\begin{equation}
    IoU_f = \frac{I_f}{U_f}, \hspace{0.5cm}
    IoU_b = \frac{I_b}{U_b}
    \label{eq:iou_fb}
\end{equation}
\begin{equation}
    FB\text{-}IoU = \frac{1}{2} (IoU_f + IoU_b)
    \label{eq:fbiou}
\end{equation}

To efficiently handle mIoU and FB-IoU in implementation, we and previous work represent $I$ and $U$ in a $2\times C$-matrix each, in which the first row stores the $I_{\neq c}$ vector and the second row the $I_{c}$ vector for the intersection matrix, likewise with $U$ for the union matrix.

\paragraph{Problem of mIoU}
In the main paper, we showed an example where a naive predictor can outperform previous work by simply predicting always foreground.
We inspect the expected performances of random prediction behaviour.
Its chance to predict a true positive in \cref{eq:ic_uc} is $r_{\hat{y}}\cdot r_{y}$, where the both terms denote the foreground ratio of the prediction and ground-truth, respectively.
The probabilities for false positives and negatives are $p(FP)=r_{\hat{y}}\cdot (1-r_{y})$ and $p(FN)=(1-r_{\hat{y}})\cdot r_{y}$, such that equation \ref{eq:iou_c} will evaluate to:
\begin{equation}
IoU_c = \frac{r_{\hat{y}} r_{y}}{r_{\hat{y}} r_{y} + r_{y} (1 - r_{\hat{y}}) + (1 - r_{y})  r_{\hat{y}}}
\label{eq:rand_iou_c}
\end{equation}
for all c, letting us obtain also $IoU_f$ in \cref{eq:iou_fb}. The background $IoU_{b}$ can be equally obtained by substituting all $r$ with $1-r$ in \cref{eq:rand_iou_c}.
From these, we can obtain both $mIoU$ and $FB\text{-}IoU$ through equations \cref{eq:miou} and \cref{eq:fbiou} respectively.
\begin{figure}[tp]
    \centering
    \includegraphics[width=\linewidth]{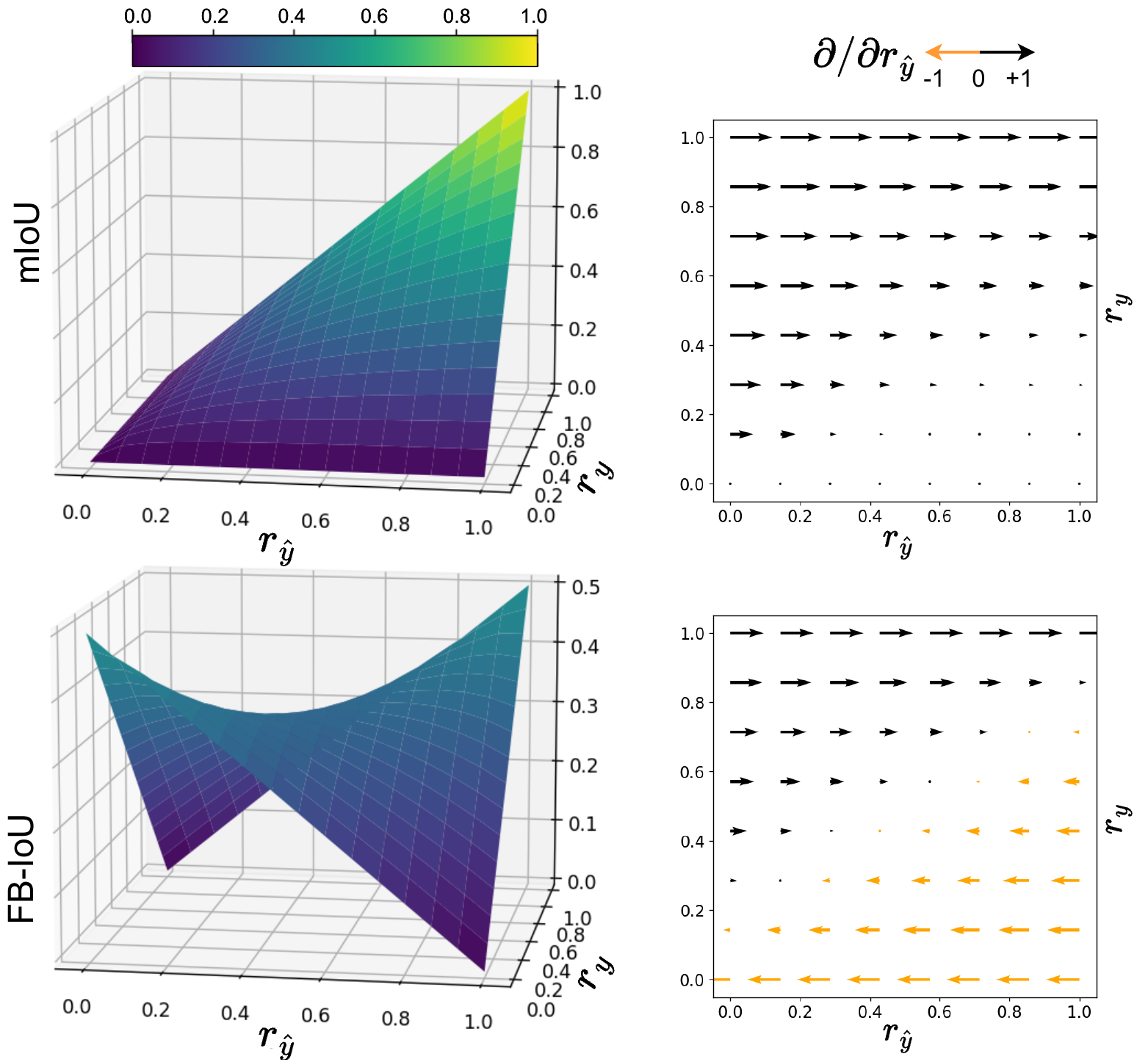}
    \caption{Results of a random mask predictor in 1-way FSS as a function of \textit{Bernoulli-sampled predicted foreground probability} $r_{\hat{y}}$ and \textit{dataset ground truth foreground ratio} $r_{y}$ (left) and its gradients with respect to the chosen predicted foreground ratio (right). For mIoU, the gradient is always positive, meaning one can get an increase in mIoU by increasing foreground prediction ratio, while for FB-IoU such overprediction is punished.}
    \label{fig:ious_rand_surface}
\end{figure}


\cref{fig:ious_rand_surface} visualizes the expected values for both metrics as a function of the foreground ratios.
The fact that a higher predicted foreground ratio leads to higher mIoU is reflected by its non-negative derivative $w.r.t.$ $r_{\hat{y}}$:
\begin{equation}
    \frac{\partial{(mIoU)}}{\partial r_{\hat{y}}} = \frac{r_{y}^{2}}{\left(r_{y} r_{\hat{y}} - r_{y} - r_{\hat{y}}\right)^{2}}
\end{equation}
In contrast, the derivative of the FB-IoU
\begin{equation}
    \frac{\partial(FB\text{-}IoU)}{\partial r_{\hat{y}}} = \frac{r_{y}^2}{(r_{\hat{y}} + r_{y} - r_{\hat{y}}r_{y})^2} - \frac{( r_{y}-1)^2}{(1 - r_{\hat{y}}r_{y})^2}
\end{equation}
can be negative and is zero at $r_{y}=r_{\hat{y}}=\frac{1}{2}$.
Compare \cref{fig:ious_rand_surface}.

\paragraph{Discussion}
We showed that mIoU performance can be boosted by increasing the foreground prediction ratio in 1-way FSS by the example of a random predictor.
In reality, the prediction has some confidence and suppressing almost-sure background naturally decreases the union area in the denominator and hence increases mIoU.
Exploiting the remaining uncertainty in a foreground-biased manner still boosts mIoU, which contrasts the intuition that the maximum performance should be reached when predicted and ground truth foreground areas match.
In standard semantic segmentation, this is less an issue, since the categories in the dataset $C$ typically equals the number of possible labels to be assigned.
However, in 1-way FSS, and in particular CD-FSS, where the uncertainty is still high, the phenomena we highlighted warrants careful consideration.
Note that the problem cannot be fixed by including the background as a semantic class for mIoU calculation, since it will still have minor contribution for large $C$.
Moreover, simply adding the background class is not semantically meaningful because the background is not a consistent class across episodes.
In 1-way episodes, there is one class selected as the foreground class, and others are treated as background.
As a consequence, background objects in one episode can be foreground objects in another.
As an alternative, we showed FB-IoU is a metric to reveal overprediction behaviour.

mIoU has been preferred over FB-IoU in previous work because it is considered to give better judgment about the generalizability of the model \citep{hs}.
This can be understood in the sense that mIoU punishes bad predictions on single classes and underrepresented classes in comparison with FB-IoU.
We agree, hence the mIoU measure should not be replaced, but complemented with the foreground ratio sensitive FB-IoU.

\begin{figure}[b]
    \centering
    \includegraphics[width=\linewidth]{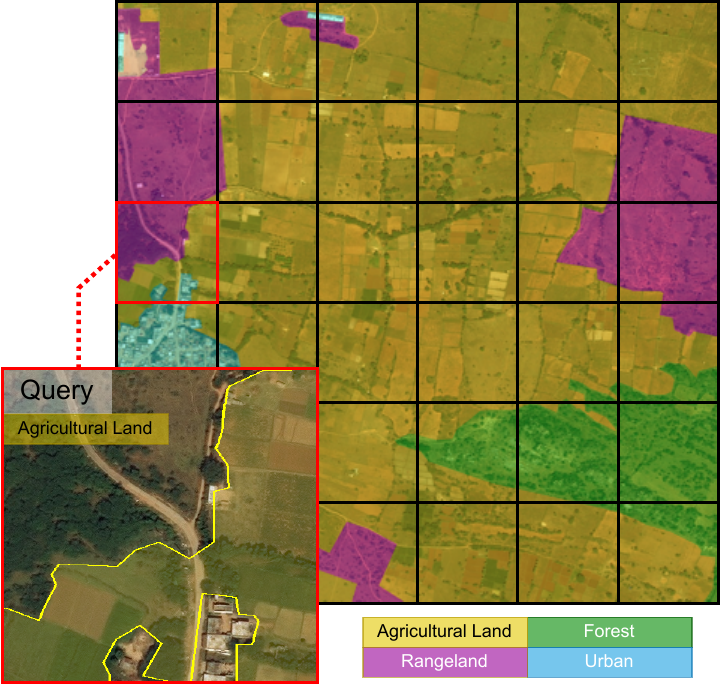}
    \caption{Cause of the Deepglobe Issue. The image from the \textit{Agricultural Land} episode we inspected in the main paper is a crop (red cell) from the here shown larger original\cite{deepglobe} image $(2448\times2448)$. Cropping is done following the CD-FSS benchmark\cite{pat}. While in the scale of the original image the inaccuracies are minor, at the zoom level of the cells it becomes intolerable. We suggest the benchmark should be adjusted accordingly. Note that also the upper left region in the query is actually \textit{Forest}, not \textit{Rangeland}.}
    \label{fig:deepglobe_grid}
\end{figure}

\begin{figure*}[bp]
  \centering
  \begin{minipage}{\textwidth}
    \centering
    \captionof{table}{Table from main paper in full. Intra- (FG$\leftrightarrow$FG) and inter- (FG$\leftrightarrow$BG) class similarities in the embedding space of (L)ow, (M)iddle and (H)igh-level feature maps. Measure represents averaged cosine similarities of pixel pairs from same and opposite classes, respectively. A higher delta represents higher discriminability.
    The intra-image statistic measures similarity within the support, across its pixel pairs which match the (FG$\leftrightarrow$FG)/(FG$\leftrightarrow$BG) criterion.
    The inter-image statistic measures similarity between query and support, across query-support pixel pairs. In case of overfitting, the intra-support discriminability would rise without bringing improvement for the inter query-support measure. The latter we argue has direct positive impact on our query-support cross-attention module, as well as the hypercorrelations in \cite{hs,pat} and dense affinity matrices in \cite{dam}.}
    \resizebox{\textwidth}{!}{%
\begin{tabular}{ccccc|ccc|ccc|ccc|ccc} 
\toprule
 && \multicolumn{3}{|c|}{Deepglobe} & \multicolumn{3}{c|}{ISIC} & \multicolumn{3}{c|}{Chest} & \multicolumn{3}{c|}{FSS} & \multicolumn{3}{c}{SUIM} \\ 
 & Metric & L & M & H & L & M & H & L & M & H & L & M & H & L & M & H \\ 
 
\midrule
\multirow{9}{*}{\rotatebox[origin=c]{90}{Before Task-Adaption}} & \multicolumn{16}{c}{Intra-Support} \\ 
\cmidrule{2-17}
 & \texttt{FG$\leftrightarrow$FG} \color{red}{\checkmark} & 0.65 & 0.52 & 0.64 & 0.73 & 0.62 & 0.73 & 0.68 & 0.57 & 0.68 & 0.56 & 0.46 & 0.64 & 0.60 & 0.54 & 0.69 \\
 & \texttt{FG$\leftrightarrow$BG} \color{teal}{\texttimes} & 0.63 & 0.47 & 0.58 & 0.70 & 0.57 & 0.63 & 0.63 & 0.48 & 0.59 & 0.51 & 0.39 & 0.56 & 0.56 & 0.45 & 0.60 \\
 & $delta$ \color{blue}{$\Delta$}  & 0.02 & 0.05 & 0.06 & 0.03 & 0.05 & 0.10 & 0.04 & 0.09 & 0.10 & 0.05 & 0.07 & 0.07 & 0.05 & 0.09 & 0.09 \\

\cmidrule{2-17}
 & \multicolumn{16}{c}{Inter-Query-Support} \\
\cmidrule{2-17}
 & \texttt{FG$\leftrightarrow$FG} \color{orange}{\checkmark} & 0.63 & 0.49 & 0.59 & 0.69 & 0.56 & 0.63 & 0.67 & 0.55 & 0.67 & 0.53 & 0.41 & 0.60 & 0.51 & 0.42 & 0.58 \\
 & \texttt{FG$\leftrightarrow$BG} \color{violet}{\texttimes} & 0.62 & 0.46 & 0.57 & 0.68 & 0.55 & 0.63 & 0.63 & 0.48 & 0.59 & 0.50 & 0.39 & 0.56 & 0.50 & 0.41 & 0.57 \\
 & $delta$ \color{cyan}{$\Delta$}  & 0.01 & 0.03 & 0.02 & 0.01 & 0.01 & 0.01 & 0.04 & 0.07 & 0.08 & 0.03 & 0.03 & 0.04 & 0.01 & 0.02 & 0.01 \\

\midrule

\multirow{9}{*}{\rotatebox[origin=c]{90}{After Task-Adaption}} & \multicolumn{16}{c}{Intra-Support} \\
\cmidrule{2-17}
 & \texttt{FG$\leftrightarrow$FG} \color{red}{\checkmark} & 0.12 & 0.26 & 0.34 & 0.19 & 0.40 & 0.53 & 0.20 & 0.36 & 0.39 & 0.29 & 0.42 & 0.47 & 0.32 & 0.45 & 0.50 \\
 & \texttt{FG$\leftrightarrow$BG} \color{teal}{\texttimes} & -0.04 & -0.11 & -0.14 & -0.06 & -0.13 & -0.16 & -0.08 & -0.14 & -0.15 & -0.11 & -0.16 & -0.16 & -0.10 & -0.14 & -0.14 \\
 & $delta$ \color{blue}{$\Delta$} & 0.17 & 0.37 & 0.48 & 0.25 & 0.53 & 0.69 & 0.28 & 0.50 & 0.54 & 0.41 & 0.58 & 0.63 & 0.42 & 0.59 & 0.65 \\

\cmidrule{2-17}
 & \multicolumn{16}{c}{Inter-Query-Support} \\
\cmidrule{2-17}
 & \texttt{FG$\leftrightarrow$FG} \color{orange}{\checkmark} & 0.03 & 0.05 & 0.05 & 0.06 & 0.13 & 0.19 & 0.17 & 0.31 & 0.33 & 0.18 & 0.28 & 0.33 & 0.14 & 0.20 & 0.16 \\
 & \texttt{FG$\leftrightarrow$BG} \color{violet}{\texttimes} & -0.01 & -0.02 & -0.01 & -0.02 & -0.04 & -0.06 & -0.06 & -0.13 & -0.12 & -0.06 & -0.10 & -0.11 & -0.04 & -0.04 & -0.03 \\
 & $delta$ \color{cyan}{$\Delta$} & 0.05 & 0.07 & 0.06 & 0.07 & 0.18 & 0.25 & 0.23 & 0.44 & 0.45 & 0.25 & 0.38 & 0.44 & 0.18 & 0.24 & 0.18 \\
\bottomrule
\end{tabular}
}

    \label{tab:embeddings_suppl}
  \end{minipage}
  \vfill 
  \begin{minipage}{\linewidth}
    \centering
    \includegraphics[width=\linewidth]{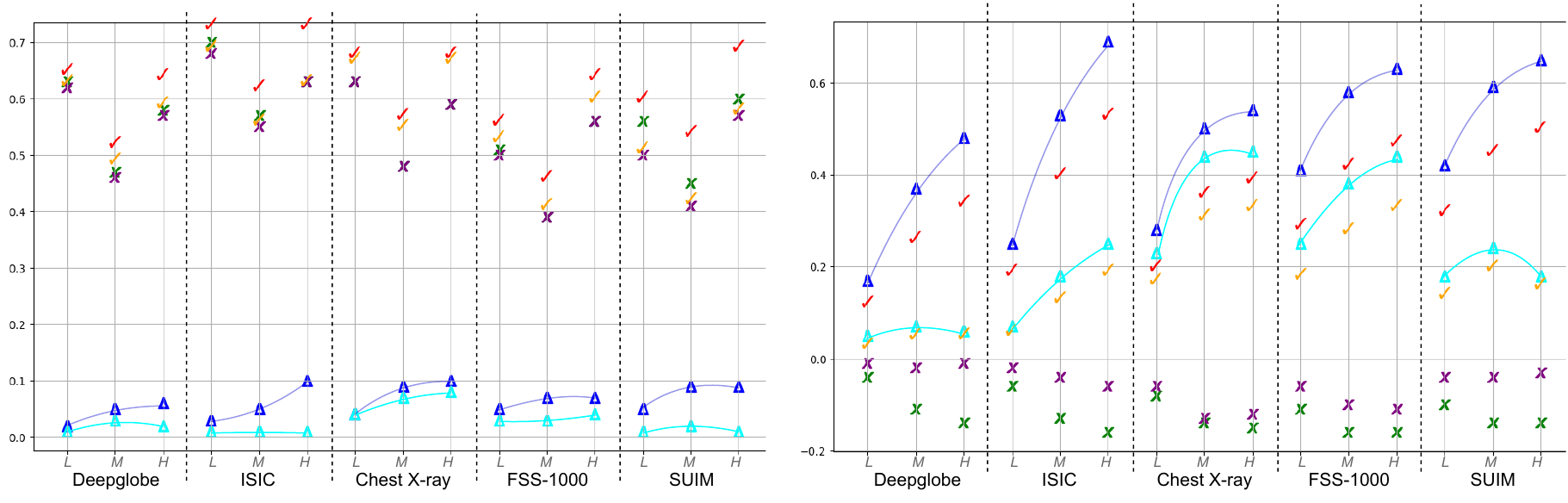}
    \captionof{figure}{Visualization of \cref{tab:embeddings_suppl}. Left: Before \textit{T}ask \textit{A}daption, right: After TA. Checkmarks represent average same-class similarity, crosses average opposite-class similarity. The most important measure for the success of the query segmentation is our discriminability measure $delta$ in cyan, representing the distance between check- and crossmarks. An overfitting to the support set could be interpreted as the vertical distance between blue and cyan in the right diagram. \textit{H}igh level features tend to be more susceptible to this (see cyan drop on Deepglobe and SUIM), but still provide important semantic information (highest on ISIC and FSS). In the main paper we noted good performance on ChestXray without TA, which is supported by seeing it to have the highest inter-query-support $delta$ in the left diagram. Note also the position of $0$ on the y-axis in both charts, indicating on the left the cyan $delta$ is almost zero for Deepglobe, ISIC and SUIM, whereas on the right TA could pushed opposite-class similarity below zero.}
    \label{fig:embeddings_suppl}
  \end{minipage}
  \label{fig:taskadaptminipage_full}
\end{figure*}

\section{Deepglobe Issue}
In the paper we argued the benchmark's\citep{pat} Deepglobe \citep{deepglobe} dataset is not appropriate due to annotation issues.
Deepglobe is an established and widely used dataset - the problem only emerges because of heavy cropping applied in the preprocessing for the benchmark.
Its creators claim that cropping has little effect because objects in sattelite images have no regular shape, but from \cref{fig:deepglobe_grid} it becomes evident that the
actual problem is that, at a higher zoom level, small spatial inaccuracies have large impact, such that almost half of the shown image is annotated wrongly.
Another example with with the same issue can be viewed in the first row of \cref{fig:qualitativecomp}.

\section{Task Adaption and Embedding Space}
\cref{tab:embeddings_suppl} reports our measures in the feature spaces after backbone and attached network respectively.
We consider this to be useful for researchers to understand the challenges in CD-FSS and our contribution to solve them.

Pixel-to-pixel similarities are measured because they are the basis for dense comparison.
We use ResNet-50 and extract the 13-layer feature pyramid following \citep{dcama,hs}.
Measurement is performed independently for each layer, their index $l$ is dropped.
Masks are first downsized by bilinear interpolation to match the feature volume size.
Intra-support similarities are obtained with the masked feature volumes
\begin{equation}
\begin{split}
&F_f^{s} = \{F^s | M^s>0.5\} \\
&F_b^{s}= F^s \setminus F_f^{s}.
\end{split}
\end{equation}
Then,
\begin{equation}
    sim_{F\leftrightarrow F}^{s\leftrightarrow s}=\frac{1}{|F_f^{s}|}^2 \sum_{f_i \in F_f^{s}} \sum_{f_j \in F_f^{s}} c(f_i, f_j)
\end{equation}
\begin{equation}
    sim_{F\leftrightarrow B}^{s\leftrightarrow s}=\frac{1}{|F_f^{s}||F_b^{s}|} \sum_{f_i \in F_f^{s}} \sum_{f_j \in F_b^{s}} c(f_i, f_j),
\end{equation}
with cosine similarity $c(\cdot)$. Equally for inter-query-support similarities, we mask the query features
\begin{equation}
    F_f^{q} = \{F^q | M^q>0.5\},
\end{equation}
\begin{equation}
    F_b^{q}= F^q \setminus F_f^{q}.
\end{equation}
Then,
\begin{equation}
    sim_{F\leftrightarrow F}^{q\leftrightarrow s}=\frac{1}{|F_f^{q}||F_f^{s}|} \sum_{f_i \in F_f^{q}} \sum_{f_j \in F_f^{s}} c(f_i, f_j)
\end{equation}
\begin{equation}
    sim_{F\leftrightarrow B}^{q\leftrightarrow s}=\frac{1}{|F_f^{q}||F_b^{s}|} \sum_{f_i \in F_f^{q}} \sum_{f_j \in F_b^{s}} c(f_i, f_j).
\end{equation}
Finally, the delta between the intra- and inter-class distances can be interpreted as the discriminability within support
\begin{equation}
    delta^{s\leftrightarrow s} = sim_{F\leftrightarrow F}^{s\leftrightarrow s} - sim_{F\leftrightarrow B}^{s\leftrightarrow s}
\end{equation}
and across (inter) query and support:
\begin{equation}
    delta^{q\leftrightarrow s} = sim_{F\leftrightarrow F}^{q\leftrightarrow s} - sim_{F\leftrightarrow B}^{q\leftrightarrow s}
\end{equation}
The block-wise $L/M/H$ measure is obtained by averaging the measure of layers belonging to a block, as in \citep{dcama,hs} the $L/M/H$ split for our 13 layers is $(4/6/3)$.

From \cref{tab:embeddings_suppl} dataset-specific characteristics become apparent.
\cref{fig:embeddings_suppl} provides an intuitive understanding of the relationship between the measures.


\section{On Affinity and Correlation Maps}
\newcommand{\timesk}{_{i=1}^{k}}
\cref{fig:qpredcoarses} visualizes the correlation maps that are the result of the dense comparison from Sec. 3.3 of the main paper.
Here we attempt to provide more intuition on their \textit{construction}, subsequent \textit{thresholding} and \textit{refinement}.
\\
\textbf{Construction} of $\hat{q}_{pred_l}$ is similar to \citep{dcama}, but since it is the core comparison mechanism of our approach, we attempt to break it down to make it more understandable why it works.
A correlation map is calculated from query features, support features and support mask.
The steps are 1) query-support pixel-to-pixel dot product, 2) softmax over the support dimension 3) filtering support foreground class.
\begin{equation}
    \hat{q}_{pred_l}=
    \underbrace{softmax(
    \overbrace{fl(\hat{\feats{q}_l}) fl(\hat{\feats{s}_l})^T
    /\sqrt{d}}
    ^{\text{1)}}
    )}
    _{\text{2)}}
    \underbrace{fl(M_l^s)}
    _{\text{3)}}.
    \label{eq:damatsuppl}
\end{equation}

1) Query features $\hat{\feats{q}_l}$ and support features $\hat{\feats{s}_l}$ are multiplied.
By flattening $fl$, feature volumes are converted into matrices with spatial dimensions represented in the first axis and channel dimensions in the second. This results in a matrix multiplication between $HW \times C$ and $C \times HW$, yielding a dense pixel-to-pixel affinity map of shape $HW\times HW$. Each element of this map is a dot product of two $C$-dimensional feature vectors, indicating the similarity between individual query and support pixels.
Division by square root of channel dimension $d$ is only scaling.

2) For any given query pixel (specific row), taking the softmax over its similarities to all support pixels (columns) accentuates support pixels with high similarity, pushing their values towards 1.

3) Multiplying a $HW$-shaped row of the affinity map with the $HW$-shaped support mask vector filters out support background regions and aggregates the remaining foreground similarities.
As a result, $\hat{q}_{pred}$ will highlight query pixels with large similarity to the support foreground.

\paragraph{Thresholding.}
Estimating the correct foreground ratio has been shown \citep{repri} to be a primary driver for performance in FSS.
We use function $thresh$ to obtain binary $\hat{M}^q$ from $\hat{q}_{fused}$.
A simple idea would be to classify every pixel with a score larger than its expected value as foreground.
For random features, the expected value of $\hat{q}_{pred}$ and thus also $\hat{q}_{fused}$ equals $mean(M^s)$, i.e. the foreground ratio in the support set, because we obtained $\hat{q}_{pred}$ by $softmax(\dotsm)M^s$ in \cref{eq:damatsuppl}.
\cref{fig:histograms} shows that the correlation scores (x-axis) are distributed around this $mean(M^s)$, but we can also observe that choosing it as a threshold would lead to overprediction.
From the shown samples it becomes apparent why a) separating the foreground cluster through k-means/Otsu's\citep{otsus} is an efficient strategy, b) we choose 
$thresh(\hat{m})=max(mean(\hat{m}), otsus(\hat{m}))$ as the threshold.
We believe the understanding of the distributions is relevant for the future development of models that want to further process correlation maps.

\begin{figure}[htbp]
    \centering
    \includegraphics[width=\linewidth]{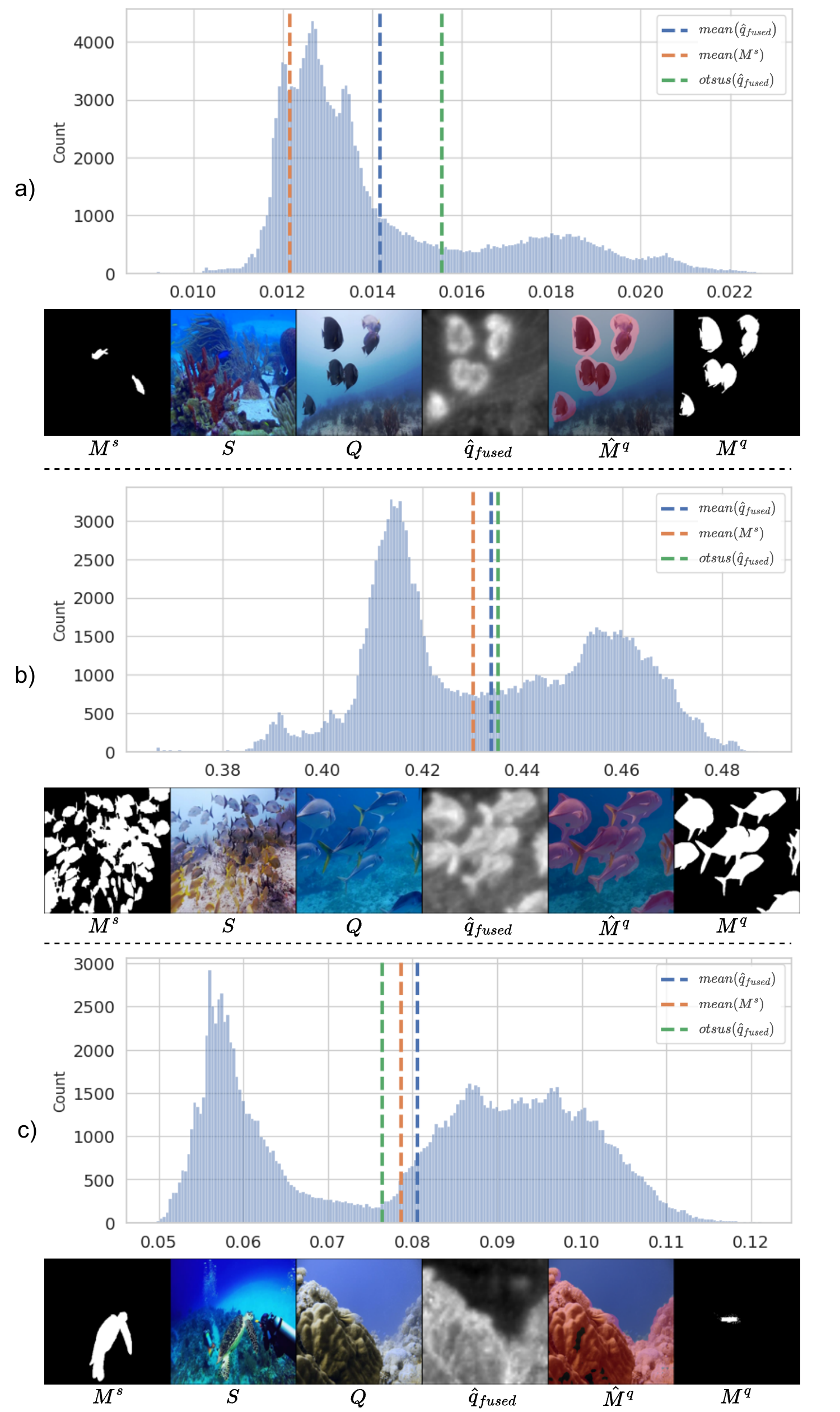}
    \caption{Histograms of correlation/prediction maps $\hat{q}_{fused}$. Cases a) and b) represent success cases where the foreground objects (right cluster in the histogram) are easily segmentable by $otsus$ (green vertical). Case c) also seems to feature two clearly distinct Gaussians, but the threshold would fall below the average prediction score across pixels (blue vertical). The right cluster is too similar to the average score, which indicates the cluster rather represents an ``unknown'' class which can be distinguished from the support background cluster (left) but is not very similar to the support foreground object. Indeed, we can see that 1) the backgrounds in $Q$ and $S$ are similar (sea), 2) the object highlighted in our $\hat{q}_{fused}$ is not similar to either support foreground (turtle) or background, 3) the actual query ground truth object (tiny hidden fish in $Q,M^q$) is visually disparate from the support turtle and hidden in unknown background, making it too difficult to segment. In this case, the average $\hat{q}_{fused}$ (blue) serves as the threshold.}
    \label{fig:histograms}
\end{figure}


\paragraph{Refinement.}
As a postprocessing step, the prediction mask $\hat{M}^q$ is refined through applying \citep{kraehenbuehl,pydensecrf}.
Not for all domains this is beneficial, and in the main paper we mentioned it can be verified by forwarding a pseudoepisode constructed from the support set.
We provide \cref{alg:postprocessing} for a detailed description of the process.
For the Chest X-ray dataset for example, it is mostly not beneficial, such that the refinement is mostly not applied.
This also reflects in Chest X-ray's slightly inverse relationship between performances \textit{Ours(no-pp)} and \textit{Ours} in Tab. 4 of the main paper.

\begin{algorithm}
\caption{Dynamic Refinement Decision.}
\begin{algorithmic}
\Require Query image $I^q$, Support set $I^s$, $M^s$ \Comment{Test Task}
\Require Orig. Support Features $\hat{\feats{s}}$
\Require Augm. Support Features $\hat{F}^{\tilde{s_1}}$ \Comment{backprojected}
\Require Prediction $\hat{q}_{fused}$ \Comment{Result of main paper Eq. 7}
\State $Q \gets \hat{\feats{s}}$ \Comment{pseudoquery}
\State $K \gets \hat{F}^{\tilde{s_1}}$ \Comment{pseudosupport}
\State $V \gets M^s$
\State $\hat{s}_{fused} \gets forward(Q,K,V)$ \Comment{main paper Eq. 6-7}
\State $\tau \gets thresh(\hat{s}_{fused})$ 
\State $\hat{M}^{s} \gets \hat{s}_{fused} > \tau$
\State $\hat{M}^{s,ref} \gets crf(I^s,\hat{s}_{fused},\tau)$ 
\If{$iou(\hat{M}^{s,ref}, M^s) > iou(\hat{M}^{s}, M^s)$}
\State $\hat{M}^q \gets crf(I^q, \hat{q}_{fused},thresh(\hat{q}_{fused}))$ \Comment{apply}
\Else
\State $\hat{M}^q \gets \hat{q}_{fused} > thresh(\hat{q}_{fused})$ \Comment{not apply}
\EndIf
\end{algorithmic}
\label{alg:postprocessing}
\end{algorithm}

Function $iou(\hat{M},M)$ calculates \cref{eq:iou_c} given prediction $\hat{M}$ and ground truth $M$.
Function $crf(I, \hat{m},\tau)$ calculates \citep{pydensecrf} with unaries from softmax generated as $sigmoid(T(\hat{m}-\tau))$, temperature T=1 for simplicity, input RGB image $I$, our soft prediction $\hat{m}$ and the calculated threshold $\tau$.

\section{Further architectural validation}
\begin{table*}
\centering
\begin{tabular}{ccccccc}
\toprule
Configuration Change & Metric & Deepglobe & ISIC & Chest-Xray & FSS-1000 & Avg. \\
\midrule
\multirow{2}{*}{a) \parbox{4cm}{\centering  kernelsize 1\textrightarrow3}} & mIoU & -8.27 & -7.90 & 2.61 & -1.60 & -3.79 \\
                   & FB-IoU & -5.57 & -14.62 & 2.01 & -2.25 & -5.11 \\
\midrule
\multirow{2}{*}{b) \parbox{4cm}{\centering out\_channels 64\textrightarrow32}} & mIoU & -0.01 & -5.28 & -0.09 & -1.60 & -1.75 \\
                   & FB-IoU & -0.20 & -5.41 & -0.20 & -1.37 & -1.80 \\
\midrule
\multirow{2}{*}{c) \parbox{4cm}{\centering out\_channels 64\textrightarrow128}} & mIoU & -0.19 & -6.25 & -0.69 & 0.16 & -1.74 \\
                   & FB-IoU & -0.11 & -5.74 & -0.39 & 0.26 & -1.50 \\
\midrule
\multirow{2}{*}{d) \parbox{4cm}{\centering n\_epochs 25\textrightarrow 10}} & mIoU & 0.21 & -6.91 & -2.69 & -0.48 & -2.47 \\
                   & FB-IoU & 0.12 & -6.36 & -1.82 & -0.10 & -2.04 \\
\midrule
\multirow{2}{*}{e) \parbox{4cm}{\centering Jitter 0\textrightarrow0.3\\ Shear 20\textrightarrow0}} & mIoU & -3.67 & -3.06 & 0.05 & -2.08 & -2.19 \\
                   & FB-IoU & -3.28 & -2.76 & 0.08 & -1.35 & -1.83 \\
\bottomrule
\end{tabular}
\caption{Performance differences under modified configurations of our attached layers. a) Replacing 1x1 convolutions through 3x3, b) Decreasing or c) increasing the number of target channels for task-adapted features, d) Fitting for less epochs, e) Replacing the augmentation method, color jitter instead of affine shearing.}
\label{tab:supplconfigs}
\end{table*}
We test our method under modified configurations.
\cref{tab:supplconfigs} reports the performance gap under these changes.
To overcome the limitations of 1x1 convolutions that do not learn relations to the spatial neighborhood, an intuitive idea would be to use a kernel size larger than 1.
However, this quadratically increases the number of learnable parameters from the sparse data and thus performs worse than the 1x1 convolutions.
We also find that geometric transformation, in our case the random shearing, is more suitable for establishing the dense consistency than operations like color jitter or blur. 

\begin{figure*}[hp]
    \centering
    \includegraphics[width=\textwidth]{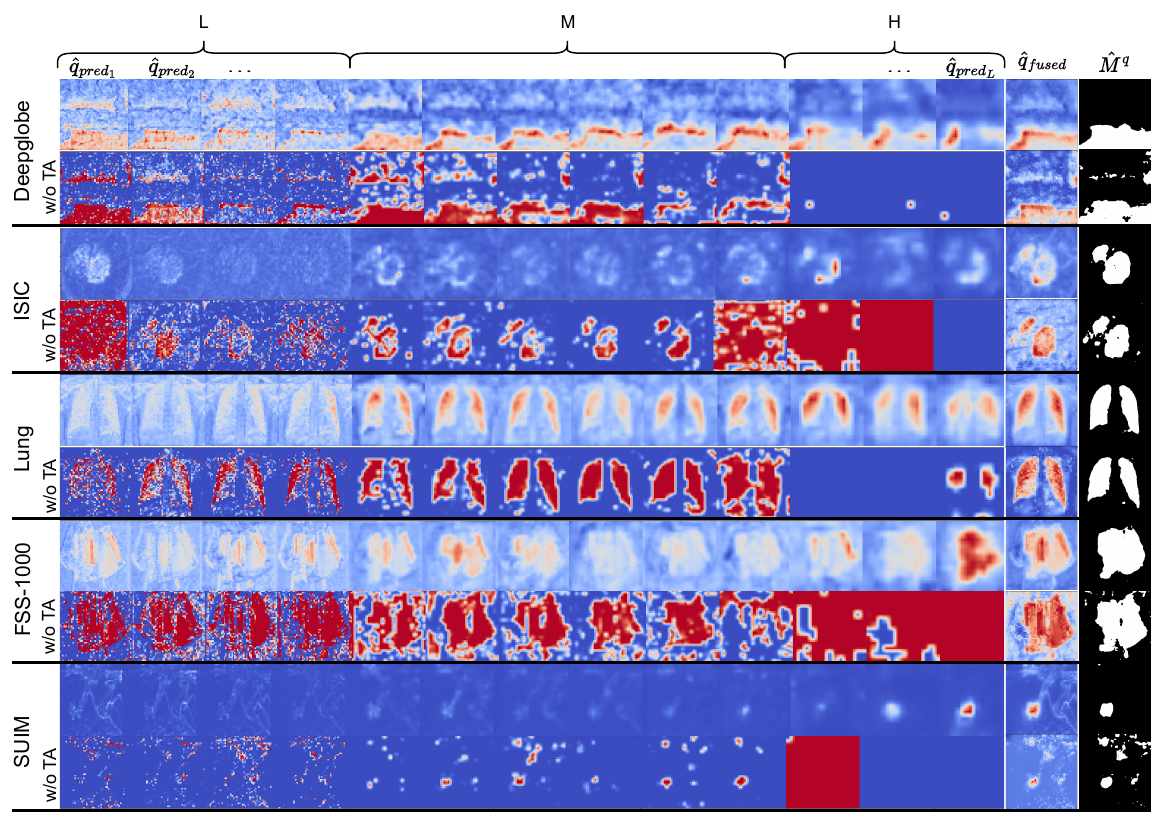}
    \caption{Layer-wise correlation maps $\hat{q}_{pred_l}$, their aggregation $\hat{q}_{fused}$ and binarization $\hat{M}^q$.
    For each dataset, the upper row represents our maps, whereas the lower row represents the results one would obtain for ResNet features, i.e. when calculating dense comparison on the feature pyramid \textit{before} our attached \textit{T}ask \textit{A}daption layers.
    Besides the improvement introduced through \textit{TA}, we can observe how considering all levels is important for CD-FSS where the target domain is unknown:
    \textit{L}ow-level features are meaningful for Deepglobe and ISIC datasets, whereas \textit{H}igh level features are more suitable for FSS and SUIM. Consistent with prior findings in FSS, mid-level layers demonstrate their utility across various datasets. Compare \cref{fig:qualitativecomp} for the sampled input images.}
    \label{fig:qpredcoarses}
\end{figure*}
\section{Qualitative Comparison}
\cref{fig:qualitativecomp} provides a qualitative comparison of our results.
The samples are the same as in \cref{fig:qpredcoarses}, such that intermediate level and final results can be compared.
\clearpage
\begin{figure*}
    \centering
    \includegraphics[width=.98\textwidth]{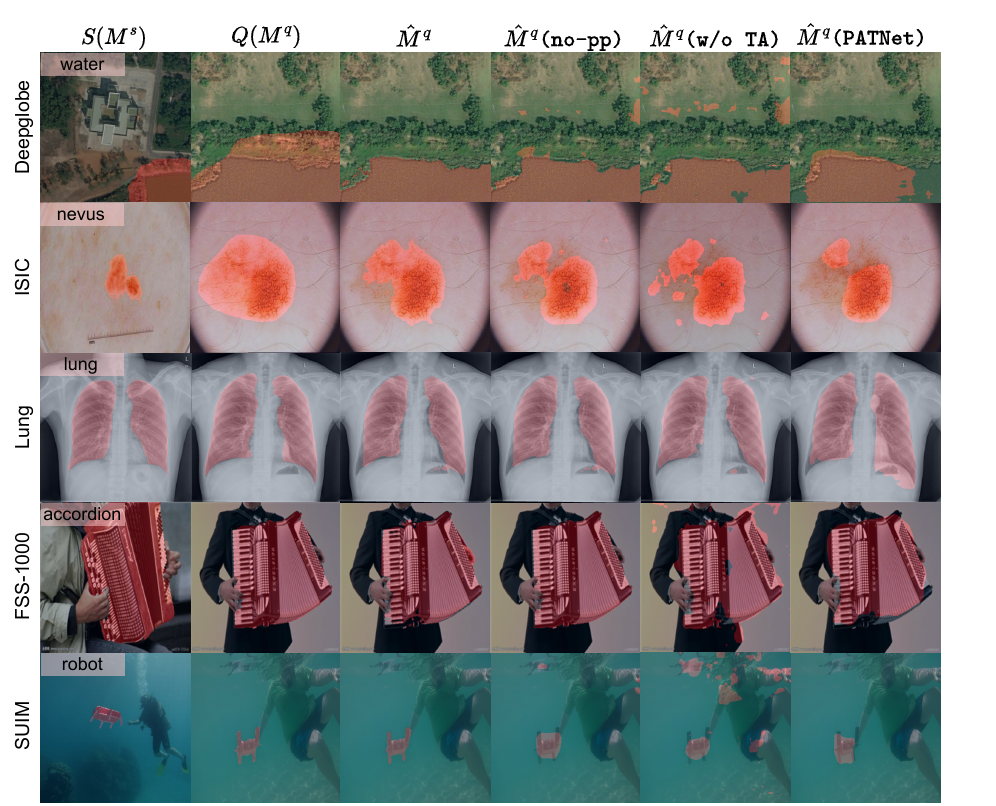}
    \caption{Qualitative comparison of results from proposed method ($\hat{M}^q$), its unrefined variant (\textit{no-pp}), ResNet feature comparison without adaption \textit{(w/o TA)} and previous CD-FSS benchmark SOTA\cite{pat} (\textit{PATNet}). We show a 1-shot episode with one \textit{S}upport image for each dataset.}
    \label{fig:qualitativecomp}
\end{figure*}

\section{Detailed Adaption and Inference Procedure}
In Sec. 4.3 of the main paper two operational modes were discussed.
On the one hand, the standard evaluation in FSS, where each test task is processed independently, without any knowledge of a previous task.
Consequently, parameters $\theta$ of attached layers $g$ are estimated from scratch for every task, corresponding to \cref{alg:adaptandfinfer}.
On the other hand, in the \textit{quick-infer} mode, parameters $\theta$ are kept constant for a task featuring a previously processed semantic class, corresponding to \cref{alg:quickinfer}.
The latter mode is useful for computational efficiency in most real-word applications where the same category should be segmented in multiple images.
For example, in the Chest-Xray dataset there is only one class, then one can run \cref{alg:adaptandfinfer} once and predict all further images only with \cref{alg:quickinfer}.
If there are tasks with different classes, then for each class parameters $\theta$ are fitted once and stored for further episodes of the same class.
\Cref{tab:linewiseruntimes} documents runtimes per operation.
In the quick-infer mode, the remaining load is primarily merely the backbone, the prediction can run at 27$|$18 fps for (1$|$5)-shot as against 15$|$3 fps of baseline\cite{huangrestnet}, shown in \Cref{tab:runtimecomparison}.
\cite{huangrestnet} is similar in architecture to PATNet\cite{pat} and HSNet\cite{hs}, but requires no test-time fine-tuning(TFI\cite{pat}) stage.
Tesla P100 was chosen for convenience, though alternative hardware may better suit local target applications.
Discrepancy between x50 factor reported in Sec. 4.3 is due to dataloader and metrics overhead.
Runtimes should be taken as initial reference only as implementations are not optimized; exemplary, improvements could be achieved by transferring thresholding to GPU, replacing the here not considered CPU-based refinement and, for the adaption, by optimizing the loss calculation.


\begin{algorithm}[t]
\caption{Adapt and infer}
\label{alg:adaptandfinfer}
\begin{algorithmic}[1] 
\Require \textcolor{teal}{ImageNet pre-trained ResNet params $\Phi$, frozen}
\Require Kaiming uniform initialized params $\{\theta_l\}_{l=1}^{L}$ of $g$
\Statex
\Require \textcolor{teal}{$Task = \{I^q, \{I_i^s,M_i^s\}\timesk\}$} \Comment{k-shot input task}
\Statex \textit{// Apply augmentation}
\State $I^{q,aug} \gets augment(I^q)$
\State \{$I_i^{s,aug}\}\timesk \gets \{augment(I_i^s)\}\timesk$
\Statex
\Statex \textit{// Forward pass through the backbone }
\State \textcolor{teal}{$F^q \gets ResNet(I^q;\Phi))$}
\State \textcolor{teal}{$\{F_i^s\}\timesk \gets \{ResNet(I_i^s;\Phi)\}\timesk$}
\State $F^{q,aug} \gets ResNet(I^{q,aug};\Phi)$
\State $\{F_i^{s,aug}\}\timesk \gets \{ResNet(I_i^{s,aug};\Phi)\}\timesk$
\Statex
\Statex \textit{// Layer-wise adaption}
\For{\textcolor{teal}{layer $l\gets 1$} \textbf{to} \textcolor{teal}{$L$}}
    \For{\textcolor{purple}{epoch $e\gets 1$} \textbf{to} \textcolor{purple}{$n_{epochs}$}}
        \Statex \hspace{3em}\textit{// Forward pass through attached layers} 
        \State \textcolor{teal}{$\hat{F}^q \gets g_l(F^q;\theta_l)$}
        \State \textcolor{teal}{$\{\hat{F}_i^s\}\timesk \gets \{g_l(F_i^s;\theta_l)\}\timesk$}
        \State $\hat{F}^{q,aug} \gets g_l(F^{q,aug};\theta_l)$
        \State $\{\hat{F}_i^{s,aug}\}\timesk \gets \{g_l(F_i^{s,aug};\theta_l)\}\timesk$
        \Statex
        \State \textcolor{purple}{Evaluate $\mathcal{L}(g_l)$ with Eq. 5 from main paper}
        \State \textcolor{purple}{Update $\theta_l$ with $SGD$: $\theta_l \gets \theta_l - \alpha\bigtriangledown_{\theta_l} \mathcal{L}(g_l)$}
    \EndFor
    \State \textcolor{teal}{$\hat{q}_{pred_l} \gets attention(\hat{F}^q,\hat{F}^s,M^s)$} \Comment{compare, Eq.6}
\EndFor
\State \textcolor{teal}{$\hat{q}_{fused} \gets \frac{1}{L}\sum_l upsample(\hat{q}_{pred_l})$ }\Comment{fuse, Eq. 7}
\State  \textcolor{teal}{$\hat{M}^q \gets thresh(\hat{q}_{fused})$} \Comment{binary pred., Eq. 8}
\end{algorithmic}
\end{algorithm}

\begin{algorithm}[t]
\caption{Infere only (quick-infer)}
\label{alg:quickinfer}
\begin{algorithmic}[1] 
\Require \textcolor{teal}{ImageNet pre-trained ResNet params $\Phi$, frozen}
\Require $\{\theta_l\}_{l=1}^{L}$ fitted by \cref{alg:adaptandfinfer}, now frozen
\Statex
\Require \textcolor{teal}{$Task = \{I^q, \{I_i^s,M_i^s\}\timesk\}$} \Comment{k-shot input task}
\Statex 
\State
\State 
\Statex
\Statex \textit{// Forward pass through the backbone}
\State \textcolor{teal}{$F^q \gets ResNet(I^q;\Phi))$}
\State \textcolor{teal}{$\{F_i^s\}\timesk \gets \{ResNet(I_i^s;\Phi)\}\timesk$}
\State 
\State 
\Statex
\Statex \textit{// Layer-wise adaption}
\For{\textcolor{teal}{layer $l\gets 1$} \textbf{to} \textcolor{teal}{$L$}}
    \State
    \Statex \hspace{1.5em}\textit{// Forward pass through attached layers} 
    \State \textcolor{teal}{$\hat{F}^q \gets g_l(F^q;\theta_l)$}
    \State \textcolor{teal}{$\{\hat{F}_i^s\}\timesk \gets \{g_l(F_i^s;\theta_l)\}\timesk$}
    \State 
    \State
    \Statex
    \State
    \State
    \State
    \State \textcolor{teal}{$\hat{q}_{pred_l} \gets attention(\hat{F}^q,\hat{F}^s,M^s)$}\Comment{compare, Eq.6}
\EndFor
\State \textcolor{teal}{$\hat{q}_{fused} \gets \frac{1}{L}\sum_l upsample(\hat{q}_{pred_l})$ }\Comment{fuse, Eq. 7}
\State \textcolor{teal}{$\hat{M}^q \gets thresh(\hat{q}_{fused})$} \Comment{binary pred., Eq. 8}
\end{algorithmic}
\end{algorithm}

\newcommand{\teal}[1]{\textcolor{teal}{#1}}
\noindent
\begin{minipage}{\linewidth}
  \centering
    \setlength{\tabcolsep}{5pt}
    \resizebox{\linewidth}{!}{%
    \newcolumntype{C}{>{\centering\arraybackslash}X}
    \begin{tabularx}{\linewidth}{ll*{8}{C}}
        \toprule
        & & \multicolumn{8}{c}{Lines} \\
        \cmidrule{3-10}
        shot & Alg. & 1+ & 3+ & 5+ & 9+ & 11+ & 13+ & 16 & 18+ \\
        \midrule
        \multirow{2}{*}{1} & \ref{alg:adaptandfinfer} & 4 & \teal{20} & 21 & \teal{1} & 1 & \textcolor{purple}{11} & \teal{1} & \teal{4} \\
        & \ref{alg:quickinfer} & - & \teal{20} & - & \teal{1} & - & - & \teal{1} & \teal{4} \\
        \bottomrule
        \\
        \toprule
        \multirow{2}{*}{5} & \ref{alg:adaptandfinfer} & 10 & \teal{35} & 63 & \teal{1} & 1 & \textcolor{purple}{16} & \teal{1} & \teal{4} \\
         & \ref{alg:quickinfer} & - & \teal{35} & - & \teal{1} & - & - & \teal{1} & \teal{4} \\
        \bottomrule
    \end{tabularx}
    }
    \captionof{table}{Runtime per line execution in pseudocode, milliseconds,\\ + indicates inclusion of the following line. Note that lines within the loop are executed more than once.}
    \label{tab:linewiseruntimes}
    \vspace{.5cm}
\end{minipage}
\begin{minipage}{\linewidth}
  \centering
     \begin{tabular}{lccc}
        \toprule
        shot & baseline\cite{huangrestnet} & Alg. \ref{alg:adaptandfinfer} & Alg. \ref{alg:quickinfer} \\\midrule
        1 & 64 & 3700 & 36 \\
        5 & 320 & 5300 & 54 \\
        \bottomrule
    \end{tabular}
    \captionof{table}{Runtime for predicting one task in milliseconds.}
    \label{tab:runtimecomparison}
\end{minipage}

In the pseudocode in \cref{alg:adaptandfinfer,alg:quickinfer}, operations printed in green are equally performed for both procedures, red lines are specific to the fitting process, equation numbers refer to the main paper, $\alpha$ is the learning rate, $L$ is the number of layers, $n_{epochs}$ is the number of iterations, all specified in the implementation details in Sec 4.1.

 \fi

\end{document}